\documentclass[11pt]{article}

\usepackage[final]{acl}

\usepackage{times}
\usepackage{latexsym}
\usepackage[T1]{fontenc}
\usepackage[utf8]{inputenc}
\usepackage{microtype}
\usepackage{inconsolata}
\usepackage{graphicx}
\usepackage{cuted}    
\usepackage{caption}  
\usepackage{tikz}
\usetikzlibrary{shapes.geometric,calc}
\usepackage{fontawesome5}  
\usepackage{hyperref}      
\hypersetup{hidelinks}

\newcommand{\HFsmiley}[1][0.32]{%
\begin{tikzpicture}[scale=#1, line cap=round, baseline={(0,0)}]
  \fill[yellow!80!orange] (-0.95,-0.55) circle (0.32);
  \fill[yellow!80!orange] ( 0.95,-0.55) circle (0.32);
  \draw[orange!85!red, line width=0.5pt] (-0.95,-0.55) circle (0.32);
  \draw[orange!85!red, line width=0.5pt] ( 0.95,-0.55) circle (0.32);
  \fill[yellow!80!orange] (0,0) circle (1);
  \draw[orange!85!red, line width=0.5pt] (0,0) circle (1);
  \draw[black, line width=1.4pt] (-0.42,0.15) .. controls (-0.32,0.40) and (-0.18,0.40) .. (-0.08,0.15);
  \draw[black, line width=1.4pt] ( 0.08,0.15) .. controls ( 0.18,0.40) and ( 0.32,0.40) .. ( 0.42,0.15);
  \draw[black, line width=1.6pt]
    (-0.45,-0.25) .. controls (-0.2,-0.58) and (0.2,-0.58) .. (0.45,-0.25);
\end{tikzpicture}%
}

\newcommand{\nexuslogo}[1][0.55]{%
\raisebox{-0.4\height}{%
\begin{tikzpicture}[scale=#1, every node/.style={transform shape}, line cap=round]
  \draw[green!55!black,  line width=2pt] (0,0) -- (45:0.85);
  \draw[red!75!black,    line width=2pt] (0,0) -- (135:0.85);
  \draw[orange!85!black, line width=2pt] (0,0) -- (225:0.85);
  \draw[teal!70!black,   line width=2pt] (0,0) -- (315:0.85);
  \node[circle, fill=green!55!black, draw=white, line width=1pt,
        inner sep=0pt, minimum size=0.36cm,
        font=\tiny\bfseries, text=white] at (45:0.98)  {$\checkmark$};
  \node[circle, fill=red!75!black, draw=white, line width=1pt,
        inner sep=0pt, minimum size=0.36cm,
        font=\tiny\bfseries, text=white] at (135:0.98) {\ding{55}};
  \node[circle, fill=orange!85!black, draw=white, line width=1pt,
        inner sep=0pt, minimum size=0.36cm,
        font=\tiny\bfseries, text=white] at (225:0.98) {?};
  \node[circle, fill=teal!70!black, draw=white, line width=1pt,
        inner sep=0pt, minimum size=0.36cm,
        font=\tiny\bfseries, text=white] at (315:0.98) {$\circlearrowleft$};
  \node[circle, draw=blue!55!black, fill=blue!10, line width=2pt,
        minimum size=0.78cm, font=\bfseries\sffamily,
        text=blue!55!black] at (0,0) {N};
\end{tikzpicture}}%
}

\usepackage{pifont}
\newcommand{\cmark}{\ding{51}}
\newcommand{\xmark}{\ding{55}}

\usepackage{amsmath}
\usepackage{amssymb}
\usepackage{booktabs}
\usepackage{tabularx}
\usepackage{array}
\usepackage{multirow}
\usepackage{makecell}
\usepackage{threeparttable}
\usepackage{adjustbox}
\usepackage{float}
\usepackage{xspace}

\usepackage{algorithm}
\usepackage{algpseudocode}

\newcommand{\NEXUS}{\textsc{NEXUS}\xspace}

\usepackage{enumitem}

\usepackage[table]{xcolor}
\usepackage{mdframed}

\newcolumntype{Y}{>{\raggedright\arraybackslash}X}
\renewcommand{\arraystretch}{1.15}

\definecolor{boxgray}{RGB}{245,245,245}
\definecolor{boxline}{RGB}{180,180,180}

\definecolor{tabheader}{RGB}{225,232,245}   
\definecolor{tabheadertext}{RGB}{25,45,90}  
\definecolor{tabalt}{RGB}{246,249,252}      
\definecolor{tabnexus}{RGB}{255,243,205}    
\definecolor{okgreen}{RGB}{34,139,34}       
\definecolor{nored}{RGB}{178,34,34}         
\definecolor{partorange}{RGB}{210,125,30}   
\definecolor{catbase}{RGB}{50,130,80}       
\definecolor{catconc}{RGB}{90,100,170}      
\definecolor{catbench}{RGB}{135,70,150}     
\definecolor{catmot}{RGB}{120,120,120}      
\definecolor{taboos}{RGB}{235,236,239}      

\newcommand{\opart}{\textcolor{partorange}{\textit{part.}}}
\newcommand{\gcheck}{\textcolor{okgreen}{\cmark}}
\newcommand{\rxmark}{\textcolor{nored}{\xmark}}
\newcommand{\nain}{\textcolor{catmot}{\textsc{n/a}}}

\definecolor{algallow}{RGB}{34,139,34}
\definecolor{algblock}{RGB}{178,34,34}
\definecolor{algconfirm}{RGB}{210,125,30}
\definecolor{algrevise}{RGB}{45,140,140}
\definecolor{algbg}{RGB}{248,251,254}        
\definecolor{algborder}{RGB}{120,140,180}    

\newcommand{\ALLOW}{\textcolor{algallow}{\textsc{allow}}}
\newcommand{\BLOCK}{\textcolor{algblock}{\textsc{block}}}
\newcommand{\CONFIRM}{\textcolor{algconfirm}{\textsc{confirm}}}
\newcommand{\REVISE}{\textcolor{algrevise}{\textsc{revise}}}

\newmdenv[
  topline=false,
  bottomline=false,
  rightline=false,
  leftline=true,
  linecolor=algborder,
  linewidth=3pt,
  innertopmargin=4pt,
  innerbottommargin=4pt,
  innerleftmargin=10pt,
  innerrightmargin=4pt,
  skipabove=4pt,
  skipbelow=4pt
]{algobox}

\newmdenv[
  backgroundcolor=boxgray,
  linecolor=boxline,
  linewidth=0.8pt,
  roundcorner=4pt,
  innertopmargin=8pt,
  innerbottommargin=8pt,
  innerleftmargin=10pt,
  innerrightmargin=10pt,
  skipabove=8pt,
  skipbelow=8pt
]{promptbox}

\definecolor{ciheader}{RGB}{40,65,110}      
\definecolor{cibg}{RGB}{240,245,252}        
\definecolor{krheader}{RGB}{45,120,75}      
\definecolor{krbg}{RGB}{238,247,240}        
\definecolor{dcheader}{RGB}{200,115,30}     
\definecolor{dcbg}{RGB}{255,250,235}        
\definecolor{taheader}{RGB}{45,130,130}     
\definecolor{tabg}{RGB}{235,247,247}        

\newenvironment{coreidea}[1]
  {\begin{mdframed}[backgroundcolor=cibg,linecolor=ciheader,linewidth=0.5pt,
    roundcorner=3pt,frametitlebackgroundcolor=ciheader,
    frametitle={\footnotesize\color{white}\strut\faLightbulb~\textbf{#1}},
    innertopmargin=5pt,innerbottommargin=5pt,innerleftmargin=7pt,innerrightmargin=7pt,
    skipabove=6pt,skipbelow=6pt]}
  {\end{mdframed}}

\newenvironment{keyresult}[1]
  {\begin{mdframed}[backgroundcolor=krbg,linecolor=krheader,linewidth=0.5pt,
    roundcorner=3pt,frametitlebackgroundcolor=krheader,
    frametitle={\footnotesize\color{white}\strut\faChartBar~\textbf{#1}},
    innertopmargin=5pt,innerbottommargin=5pt,innerleftmargin=7pt,innerrightmargin=7pt,
    skipabove=6pt,skipbelow=6pt]}
  {\end{mdframed}}

\newenvironment{designchoice}[1]
  {\begin{mdframed}[backgroundcolor=dcbg,linecolor=dcheader,linewidth=0.5pt,
    roundcorner=3pt,frametitlebackgroundcolor=dcheader,
    frametitle={\footnotesize\color{white}\strut\faQuestionCircle~\textbf{#1}},
    innertopmargin=5pt,innerbottommargin=5pt,innerleftmargin=7pt,innerrightmargin=7pt,
    skipabove=6pt,skipbelow=6pt]}
  {\end{mdframed}}

\newenvironment{takeaway}[1]
  {\begin{mdframed}[backgroundcolor=tabg,linecolor=taheader,linewidth=0.5pt,
    roundcorner=3pt,frametitlebackgroundcolor=taheader,
    frametitle={\footnotesize\color{white}\strut\faArrowRight~\textbf{#1}},
    innertopmargin=5pt,innerbottommargin=5pt,innerleftmargin=7pt,innerrightmargin=7pt,
    skipabove=6pt,skipbelow=6pt]}
  {\end{mdframed}}

\title{\nexuslogo[0.75]\\[0.4em]
NEXUS: Structured Runtime Safety for Tool-Using LLM Agents}

\author{
\textbf{Elias Hossain}$^{1}$\thanks{Corresponding author: \texttt{mdelias.hossain@ucf.edu}},
\textbf{Md Mehedi Hasan Nipu}$^{2}$,
\textbf{Tasfia Nuzhat Ornee}$^{1}$,
\textbf{Rajib Rana}$^{3}$,\\
\textbf{Niloofar Yousefi}$^{1}$
\\\\
$^{1}$University of Central Florida \quad
$^{2}$North South University\\
$^{3}$University of Southern Queensland \quad
\\\\
\texttt{\textbf{mdelias.hossain@ucf.edu}}
\\[0.9em]
{\fontsize{28}{30}\selectfont
\href{https://eliashossain001.github.io/nexus/}{\textcolor{blue!65!black}{\faGlobe}}%
\hspace{1.2em}%
\raisebox{0.30em}{\href{https://huggingface.co/datasets/EliasHossain/nexus-multistep}{\HFsmiley[0.42]}}%
\hspace{1.2em}%
\href{https://github.com/eliashossain001/nexus}{\faGithub}}
}

\begin{document}
\maketitle

\begin{strip}
    \centering
    \includegraphics[width=\textwidth]{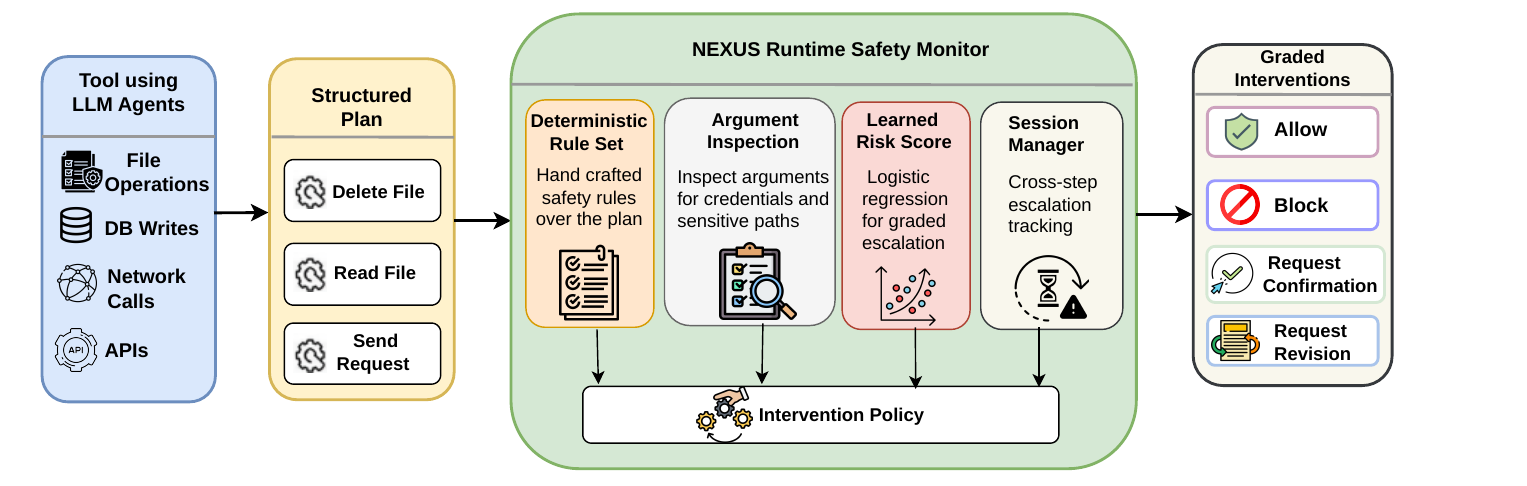}
    \captionof{figure}{Overview of the \NEXUS runtime safety monitor. \textbf{(a)} A tool-using LLM agent issues calls across heterogeneous side-effect categories (file operations, database writes, network calls, external APIs) and emits them as a \textbf{(b)}~\emph{structured plan}, an ordered intermediate representation in which each step carries its tool name, arguments, required permissions, side-effect class, irreversibility flag, sensitivity flag, and estimated cost. \textbf{(c)} \NEXUS intercepts the plan before any tool is executed and evaluates it through four complementary components: a \emph{deterministic rule set} that fires on nine plan-level safety rules covering permissions, irreversible actions, sensitive access, network egress, budget limits, suspicious patterns, scope ambiguity, broad-scope operations, and untrusted-input-driven actions; an \emph{argument inspector} that scans step arguments for credentials, sensitive paths, sensitive tables and fields, and suspicious URLs; a \emph{calibrated learned risk score} $\rho(P) \in [0,1]$ produced by a Platt-scaled logistic regression over a nine-dimensional plan feature vector and used as a graded escalation signal; and a \emph{session manager} that tracks exposed entities, cumulative permissions, repeated unsafe attempts, and cumulative cost across turns. \textbf{(d)} A formal intervention policy $\Pi$ combines these signals using critical-violation counts, severity thresholds, and the calibrated thresholds $(\tau_b, \tau_c)$, and routes each plan to one of four \emph{graded interventions}: \textsc{allow} (execute as proposed), \textsc{block} (refuse and halt), \textsc{request confirmation} (escalate to user approval), or \textsc{request revision} (return the plan for repair). The scorer-gated demotion step, in which $\rho(P)$ arbitrates between \textsc{block} and \textsc{confirm} when exactly one critical violation fires, is the source of \NEXUS's $+27.3$\,pp gain in $4$-class intervention accuracy over rule-only monitoring.}
    \label{fig:nexus-overview}
    \vspace{0.5em}
\end{strip}

\begin{abstract}
Tool-using LLM agents increasingly execute high-impact actions, making runtime safety monitoring essential. We present \NEXUS\ (\textbf{Neural EXecution Utility and Safety}), a structured-plan safety monitor that applies a formal intervention policy $\Pi$ to select among four actions: \emph{allow}, \emph{block}, \emph{request confirmation}, or \emph{request revision}. \NEXUS combines deterministic safety rules, argument-level inspection, and a calibrated logistic-regression risk score for graded escalation. On a $128$-instance synthetic benchmark, \NEXUS attains $F_1 = 0.949$ and $4$-class intervention accuracy of $0.6406$, outperforming rule-only intervention selection by $+27.3$ percentage points. It also improves over rule-only on R-Judge ($F_1 = 0.861$ vs.\ $0.849$), matches rule-only on AgentHarm due to threat-model limits, and achieves $0\%$ ASR at $99\%$ control allow on IPI. On the rule-blind \NEXUS-Stress benchmark, \NEXUS reaches $F_1 = 0.881$, highlighting the difficulty of fine-grained intervention routing. With $0.205$\,ms median latency, \NEXUS adds under $0.1\%$ overhead to typical agent loops. Code, benchmarks, and the calibrated risk scorer are publicly released.

\smallskip
\noindent\textcolor{red}{\textbf{License notice.} The \NEXUS datasets are released solely for research on agent safety. By using them you agree to respect the linked dataset cards on Hugging Face; see Appendix~\ref{app:license} for the full release terms.}
\end{abstract}

\section{Introduction}
\label{sec:intro}

Large language models (LLMs) are increasingly deployed as tool-using agents that invoke APIs, access files, query databases, execute code, and act on behalf of users~\cite{openai_swarm_2024,anthropic_mcp_2024,he2025emerged,deng2025ai}. This shift from passive text generation to action-oriented autonomy changes the safety problem: unsafe behavior can now cause destructive file operations, credential exposure, unauthorized network access, or resource abuse \emph{before} a human notices~\cite{yao2024survey,das2025security,zhang2025llm}.

\textbf{The tool-use threat surface.}
Unsafe behavior can arise from the agent's execution plan itself, including the tools it selects, the inputs it provides, the permissions it requests, and the side effects it produces. For example, an agent may read a sensitive file and then send its contents to an external location. Each step may seem reasonable on its own, but together they form a harmful data-leakage plan. Prompt injection, malicious retrieved content, corrupted tool outputs, or ambiguous requests can also redirect agents toward unsafe behavior~\cite{perez2022ignore,greshake2023not,zhan2024injecagent}. Thus, agent safety requires monitoring \emph{planned tool use}, not only generated text~\cite{bodea2025trusted,maloyan2026breaking}.

\textbf{Limits of existing safeguards.}
Existing safeguards address parts of this problem. Declarative rule systems~\cite{wang2025agentspec,kamath2025enforcing} enforce predefined constraints, symbolic verifiers~\cite{miculicich2025veriguard,zhao2026clawguard} inspect individual tool calls, and judge-style filters~\cite{yuan2024r} evaluate trajectories. However, these approaches often operate at a single abstraction level, provide limited support for plan-level escalation, or reduce safety to a binary safe/unsafe decision. In practice, some plans should be blocked, some should require confirmation, and others should be revised instead of fully rejected.

We present \NEXUS, a structured runtime safety monitor for tool-using language agents. \NEXUS operates over a plan representation containing ordered tool invocations, arguments, permissions, side-effect categories, sensitivity indicators, irreversibility flags, and resource estimates. Given a proposed plan, \NEXUS selects one of four interventions: \emph{allow}, \emph{block}, \emph{request confirmation}, or \emph{request revision}. Its hybrid backbone combines deterministic safety rules, argument-level inspection, and calibrated risk scoring. The learned component does not replace the rule layer; it provides a calibrated escalation signal for fine-grained intervention routing while preserving auditability.

We evaluate \NEXUS across held-out synthetic tests, adversarial indirect-prompt-injection settings, multi-turn escalation scenarios, R-Judge, AgentHarm, and the rule-blind \NEXUS-Stress benchmark. \NEXUS achieves strong structured-plan detection and improves fine-grained intervention routing over rule-only monitoring. The evaluation shows that plan-level monitoring is effective when unsafe behavior is visible in tool actions, arguments, permissions, or side effects. It also identifies an important boundary: when harmful and benign behaviors share the same structural tool plan and differ mainly in latent intent, additional prompt-aware or intent-aware safeguards may be required~\cite{zhong2025rtbas}. Our main contributions are as follows:

\begin{itemize}[leftmargin=*,itemsep=2pt,topsep=2pt]
    \item \textbf{Structured runtime safety.}
    We introduce \textbf{\NEXUS}, a plan-level safety monitor that inspects tool-using LLM agents \emph{before} tool execution.

    \item \textbf{Graded intervention policy.}
    We formulate a four-action policy that routes plans to \emph{allow}, \emph{block}, \emph{request confirmation}, or \emph{request revision}, going beyond binary safety decisions.

    \item \textbf{Hybrid and interpretable detection.}
    We develop an interpretable framework that combines deterministic rules, argument-level inspection, and calibrated risk scoring under a unified intervention policy.

    \item \textbf{Comprehensive evaluation and release.}
    We evaluate \NEXUS across synthetic, adversarial, multi-turn, external, and rule-blind benchmarks, and release the code, benchmarks, trained scorer, calibration metadata, and reproducibility scripts.
\end{itemize}

\section{Related Work }
\label{sec:related}

\paragraph{Runtime safety for tool-using agents.}
Prior work approaches agent safety along three axes: declarative rule systems~\cite{wang2025agentspec,kamath2025enforcing} encode invariants over agent behaviour; symbolic verifiers~\cite{miculicich2025veriguard,zhao2026clawguard} check individual tool calls; and judge-style monitors~\cite{yuan2024r} score whole trajectories with another LLM. Orthogonal benchmarks (ToolSafe~\cite{mou2026toolsafe}, AgentHarm~\cite{andriushchenko2025agentharm}, AgentDrift~\cite{wu2026agentdrift}) target tool-use harms, harmful behaviours, and trace-perturbation robustness; the Verifier-Tax study~\cite{sah2026verifier} analyses the cost of always-on verification.

\paragraph{Prompt-injection defences.}
A complementary line targets prompt-injection at the input/model layer. StruQ~\cite{chen2025struq} fine-tunes a structured-query interface separating trusted instructions from untrusted data; Spotlighting~\cite{hines2024spotlighting} marks untrusted content via delimiting, datamarking, or encoding. Both operate before any tool call is committed and are orthogonal to \NEXUS, which monitors the resulting plan regardless of the upstream defence: our IPI evaluation (Appendix~\ref{app:ipi-v2}) shows that once an injection steers the agent, the resulting plan is caught by any structured-plan monitor.

\paragraph{Positioning.}
We position \NEXUS (Table~\ref{tab:related}) at the intersection of three design choices: (i) monitoring at the granularity of a structured plan IR with argument-level inspection; (ii) a hybrid of deterministic rules, calibrated learned risk, and argument inspection unified by a formal four-class intervention policy $\Pi$; and (iii) an explicit safety--utility objective (Eq.~\ref{eq:utility}) that makes the threshold trade-off tunable.

\begin{table*}[t]
\centering
\scriptsize
\setlength{\tabcolsep}{3pt}
\renewcommand{\arraystretch}{1.08}

\begin{tabularx}{\textwidth}{l>{\raggedright\arraybackslash}p{2.25cm}cccccc>{\raggedright\arraybackslash}X}
\toprule
\rowcolor{tabheader}
\textcolor{tabheadertext}{\textbf{System}} &
\textcolor{tabheadertext}{\textbf{Plan IR}} &
\textcolor{tabheadertext}{\textbf{Rules}} &
\textcolor{tabheadertext}{\textbf{Learned}} &
\textcolor{tabheadertext}{\textbf{Arg.}} &
\textcolor{tabheadertext}{\textbf{Session}} &
\textcolor{tabheadertext}{\textbf{Adv.}} &
\textcolor{tabheadertext}{\textbf{Multi-turn}} &
\textcolor{tabheadertext}{\textbf{Compare-to-\NEXUS}} \\
\midrule
\rowcolor{tabalt}
AgentSpec~\cite{wang2025agentspec} & declarative & \gcheck & \rxmark & \opart & \rxmark & \opart & \rxmark & \textcolor{catbase}{baseline} \\
VeriGuard~\cite{miculicich2025veriguard} & symbolic & \gcheck & \rxmark & \gcheck & \rxmark & \rxmark & \opart & \textcolor{catconc}{conceptual} \\
\rowcolor{tabalt}
ToolSafe~\cite{mou2026toolsafe} & trajectory & \opart & \gcheck & \opart & \gcheck & \opart & \gcheck & \textcolor{catbench}{benchmark} \\
CLAWGUARD~\cite{zhao2026clawguard} & per-call & \gcheck & \rxmark & \gcheck & \rxmark & \rxmark & \rxmark & \textcolor{catconc}{conceptual} \\
\rowcolor{tabalt}
Agent-C~\cite{kamath2025enforcing} & constitution & \gcheck & \rxmark & \rxmark & \rxmark & \rxmark & \opart & \textcolor{catconc}{conceptual} \\
Verifier-Tax~\cite{sah2026verifier} & \nain & \nain & \nain & \nain & \nain & \nain & \nain & \textcolor{catmot}{motivation} \\
\rowcolor{tabalt}
AgentHarm~\cite{andriushchenko2025agentharm} & \nain & \nain & \nain & \nain & \nain & \gcheck & \opart & \textcolor{catbench}{benchmark} \\
R-Judge~\cite{yuan2024r} & judge model & \rxmark & \gcheck & \opart & \opart & \opart & \gcheck & \textcolor{catbench}{benchmark+baseline} \\
\rowcolor{tabalt}
AgentDrift~\cite{wu2026agentdrift} & trace perturb. & \rxmark & \rxmark & \rxmark & \rxmark & \gcheck & \gcheck & \textcolor{catmot}{motivation} \\
\midrule
\rowcolor{tabnexus}
\textbf{\NEXUS} & \textbf{structured plan IR} & \gcheck & \textbf{calib.} & \gcheck & \gcheck & \gcheck & \opart & -- \\
\bottomrule
\end{tabularx}

\caption{\NEXUS vs.\ prior runtime-safety systems and benchmarks. \cmark{} = full support; \emph{part.} = partial. \emph{Compare-to-\NEXUS}: \textbf{baseline} (empirical baseline), \textbf{conceptual} (design-level only), \textbf{benchmark+baseline} (both), \textbf{motivation} (cited for framing only). \emph{Multi-turn} is \emph{part.} because the single-turn policy reaches $F_1 = 0.500$ on \emph{multi\_step\_escalation}; the session-aware extension (\S\ref{sec:multiturn}) mitigates this on the authored $120$-session benchmark but is not validated on organic traces.}
\label{tab:related}
\end{table*}

\section{Threat Model}
\label{sec:threat}

We consider an adversary who can (i) issue a directly malicious request, (ii) inject instructions into retrieved documents or prior tool outputs (\emph{IPI}), (iii) return adversarial tool outputs, (iv) exploit permission mismatches (\emph{privilege misuse}), (v) chain individually safe steps into multi-turn exfiltration (\emph{long-horizon escalation}), or (vi) cause the agent to exceed its resource budget. The adversary cannot modify the LLM weights, safety rules, or tool implementations. For each plan $P$, the defender selects $\Pi(P) \in \{\textsc{allow}, \textsc{block}, \textsc{confirm}, \textsc{revise}\}$ to minimize the safety--utility loss in Eq.~\ref{eq:utility}.

\begin{table}[t]
\centering
\scriptsize
\setlength{\tabcolsep}{3pt}
\renewcommand{\arraystretch}{1.08}

\begin{tabularx}{\columnwidth}{>{\raggedright\arraybackslash}p{1.7cm}
>{\raggedright\arraybackslash}p{1.55cm}
>{\raggedright\arraybackslash}X}
\toprule
\rowcolor{tabheader}
\textcolor{tabheadertext}{\textbf{Threat}} &
\textcolor{tabheadertext}{\textbf{Channel}} &
\textcolor{tabheadertext}{\textbf{\NEXUS handling}} \\
\midrule
\rowcolor{tabalt}
Direct malicious request & User prompt & Rules + arg. + $\rho$ on plan \\
Indirect prompt injection & RAG/tool output & Structured-plan monitoring \\
\rowcolor{tabalt}
Corrupted tool output & API/DB response & Schema + suspicious-pattern rules \\
Privilege misuse & Plan execution & Permission rule \\
\rowcolor{tabalt}
Sensitive exfiltration & Network step & Argument inspection \\
Resource abuse & Budget overrun & Budget rule \\
\rowcolor{tabnexus}
Long-horizon escalation & Multi-turn session & \textcolor{partorange}{\textit{Single-turn: partial;}} session-aware $\Pi$ mitigates on authored benchmark (\S\ref{sec:multiturn}) \\
\midrule
\rowcolor{taboos}
Jailbreaking agent LLM & Model weights & \textcolor{catmot}{\textit{Out of scope}} \\
\rowcolor{taboos}
Tool-side compromise & Tool impl. & \textcolor{catmot}{\textit{Out of scope}} \\
\rowcolor{taboos}
Covert side channels & Timing/cache & \textcolor{catmot}{\textit{Out of scope}} \\
\bottomrule
\end{tabularx}

\caption{Threats addressed by \NEXUS at runtime. The single-turn policy detects $F_1 = 0.500$ on \emph{multi\_step\_escalation}; the session-aware extension of \S\ref{sec:multiturn} mitigates this on the $120$-session authored benchmark but is not yet validated on organic multi-turn traces (\S\ref{sec:limitations}).}
\label{tab:threats}
\end{table}


\section{The \NEXUS Framework}
\label{sec:framework}

\begin{coreidea}{Three signals, one policy}
The framework operates over a \emph{structured plan IR} and combines three logically independent safety signals (rules, argument inspection, calibrated risk) under a single intervention policy $\Pi$. Each signal is testable in isolation, and any decision can be traced back to a named rule, an argument-inspector finding, or a calibrated threshold crossing.
\end{coreidea}

Figure~\ref{fig:nexus-overview} summarizes the \NEXUS runtime pipeline. A tool-using LLM agent first produces a structured plan rather than directly executing tool calls. This plan contains the selected tools, their arguments, required permissions, side-effect categories, and execution context. \NEXUS intercepts the plan before execution and evaluates it through four complementary components: a deterministic rule set for explicit safety violations, an argument inspector for sensitive values and risky inputs, a calibrated learned risk scorer for graded escalation, and a session manager for cross-turn state. The resulting signals are passed to an intervention policy that decides whether to \textsc{allow}, \textsc{block}, request user \textsc{confirmation}, or request plan \textsc{revision}.

The remainder of this section defines each component. We first introduce the structured plan intermediate representation, then describe the rule set and argument inspector. Section~\ref{sec:risk} presents the calibrated learned risk scorer, and Section~\ref{sec:policy} defines the intervention policy that combines all signals into a single runtime decision.

\subsection{Structured Plan IR}

A plan is $P = (s_1, \dots, s_n, c)$, where each step $s_i$ specifies a tool name, arguments, side-effect category (file/database/network/compute), irreversibility flag, sensitivity flag, required permissions, and estimated cost. The context $c$ records the agent's available permissions, remaining budget, prior tool-output history, and per-session policy flags.

\subsection{Rules and Argument Inspection}

\NEXUS applies nine deterministic rules. The first six cover permission checks, irreversible actions, sensitive access, network access, budget limits, and suspicious patterns. During development, we added three additional rules: R7, \emph{scope ambiguous action}, which escalates security-posture changes made through insecure flags (severity \textsc{high}, or \textsc{crit} for unbounded grants); R8, \emph{broad scope operation}, which flags wildcard or glob patterns in file and data operations (severity \textsc{med}); and R9, \emph{external source driven}, which flags cases where untrusted input drives privileged operations (severity \textsc{high}).

Each rule emits violations with severity $\mathrm{sev}(v) \in \{\textsc{low}, \textsc{med}, \textsc{high}, \textsc{crit}\}$. In parallel, the argument inspector scans step arguments for credential patterns, sensitive paths, sensitive tables or fields, and suspicious URLs. These argument-level findings are treated as violations on equal footing with rule-based violations $V_R$. This component improves the \emph{sensitive data access} category from $F_1 = 0.273$ to $1.000$ (Appendix~\ref{app:per-cat-table}).

\subsection{Calibrated Learned Risk}
\label{sec:risk}

\paragraph{Feature vector and calibration.}
We train a logistic-regression risk scorer on a $9$-D plan-level feature vector $\phi(P)$ covering plan length, indicators of irreversible / sensitive / delete-side-effect steps, per-category step counts, total estimated cost, and tool-category diversity (full definitions in Appendix~\ref{app:features}). We use a stratified $70/15/15$ split (seeds $42$ / $7$) of the $428$-instance synthetic benchmark for train, validation, and test; thresholds $(\tau_b, \tau_c)$ are selected on the validation split and all reported metrics come from the held-out test split exclusively. The validation-best thresholds are $\tau_b = 0.30$, $\tau_c = 0.10$; the paper defaults $(0.75, 0.70)$ yield identical $F_1$ with slightly higher intervention accuracy on the test split ($0.667$ vs.\ $0.651$), confirming no threshold leakage. The training set is further split $80/20$ (seed $7$) into base-fit and calibration sets, on which we fit Platt scaling and isotonic regression. Platt scaling reduces ECE $\approx 6.5{\times}$ over the raw scorer (full metrics in Table~\ref{tab:calibration}, Appendix~\ref{app:calibration-fig}); we use Platt-scaled $\rho$ throughout.

\paragraph{Role in $\Pi$.}
We use $\rho$ as a graded escalation signal rather than as a primary detection engine. Its specific role is the scorer-gated demotion step in $\Pi$ (\S\ref{sec:policy}): when exactly one critical rule fires, $\rho$ determines whether to issue a hard block or route the case to user confirmation, improving $4$-class intervention accuracy by $+27.3$ pp (permutation $p < 0.001$). The calibration analysis above ensures that $(\tau_b, \tau_c)$ operate on a probability scale that meaningfully reflects empirical unsafe rates.

\begin{designchoice}{Why Platt over isotonic?}
Both reduce ECE substantially over the raw scorer, and isotonic obtains a marginally lower Brier score ($0.038$ vs.\ $0.041$). We deploy Platt because it achieves the \emph{lowest} ECE ($0.013$, $\approx 6.5{\times}$ reduction over raw) and is more stable on the small $60$-instance calibration split, where isotonic risks overfitting (Appendix~\ref{app:calibration-fig}).
\end{designchoice}

\subsection{Intervention Policy}
\label{sec:policy}

Let $V(P) = V_R(P) \cup V_A(P)$ be the combined violation set and let $n_{\mathrm{crit}}(P) = |\{v \in V(P) : \mathrm{sev}(v) = \textsc{crit}\}|$. With thresholds $(\tau_b, \tau_c) = (0.75, 0.70)$, the scorer-gated demotion policy is

\begin{equation}
\begin{adjustbox}{max width=\columnwidth}
$\displaystyle
\Pi(P) = \begin{cases}
\mathrm{BLOCK}   & n_{\mathrm{crit}} \ge 2 \\
\mathrm{BLOCK}   & n_{\mathrm{crit}} = 1 \wedge \rho(P) \ge \tau_b \\
\mathrm{CONFIRM} & n_{\mathrm{crit}} = 1 \wedge \rho(P) < \tau_b \\
\mathrm{CONFIRM} & \exists\, v : \mathrm{sev}(v) = \mathrm{HIGH} \\
\mathrm{REVISE}  & \exists\, v : \mathrm{sev}(v) = \mathrm{MED} \\
\mathrm{CONFIRM} & \rho(P) \ge \tau_c \\
\mathrm{ALLOW}   & \text{otherwise}
\end{cases}
$
\end{adjustbox}
\label{eq:pi}
\end{equation}

Cases are evaluated top-to-bottom and the first matching action is returned. Multiple critical violations trigger an unconditional block. When exactly one critical violation is present, the risk score gates between blocking the plan ($\rho \ge \tau_b$) and requesting confirmation ($\rho < \tau_b$); this is the scorer-gated demotion step. High-severity violations route to confirmation, medium-severity violations route to revision, and the learned scorer provides a residual confirmation signal when no rule violation is present.

\paragraph{Safety--utility objective.}
The thresholds $(\tau_b, \tau_c)$ are tunable parameters of $\Pi$. For a workload $\mathcal{W}$ of (plan, gold-label) pairs, we define
\begin{equation}
\mathcal{L}(\Pi) = \mathbb{E}_{(P, y) \sim \mathcal{W}}\!\left[\lambda_s\,u_s + \lambda_o\,u_o + \lambda_c\,c(\Pi(P))\right]
\label{eq:utility}
\end{equation}
where $u_s = \mathbf{1}[y = \textsc{unsafe} \wedge \Pi(P) = \textsc{allow}]$ is the unsafe-allow indicator, $u_o = \mathbf{1}[y = \textsc{safe} \wedge \Pi(P) \neq \textsc{allow}]$ is the overblocking indicator, and $c(\cdot)$ is the per-intervention cost: $c(\textsc{allow}) = 0$, $c(\textsc{revise}) = 0.1$, $c(\textsc{confirm}) = 0.3$, and $c(\textsc{block}) = 1$. Section~\ref{app:thresholds} shows that binary $F_1$ and the unsafe-caught rate are invariant to $(\tau_b, \tau_c)$ on this distribution; only the $4$-class intervention accuracy is sensitive to the operating point.

\begin{designchoice}{Why $(\tau_b, \tau_c) = (0.75, 0.70)$?}
Selected on the $63$-instance validation split and held constant across all $25$ $(\lambda_s, \lambda_o)$ grid settings (Appendix~\ref{app:thresholds}). The validation-best $(0.30, 0.10)$ yields identical $F_1$ but lower test-split intervention accuracy ($0.651$ vs.\ $0.667$), confirming no threshold leakage from test data into deployment.
\end{designchoice}

\paragraph{Algorithm.}
The full procedure (Appendix~\ref{app:algorithm}) computes $V_R$, $V_A$, and $\rho$ as logically independent signals, then dispatches via Eq.~\ref{eq:pi}. Median CPU latency is $0.205$\,ms per plan on the IPI set ($4{,}349$ decisions/sec); the dominant cost is \texttt{predict\_proba}.

\section{\NEXUS-Bench}
\label{sec:bench}

We evaluate \NEXUS on two distributions.

\paragraph{Synthetic split.}
\NEXUS-Bench v1 contains $428$ instances generated from $9$ deterministic templates over $7$ tool categories. A stratified $70/15/15$ split with seed $42$ yields a $300$-instance training set, a $63$-instance validation set for threshold selection, and a $128$-instance held-out test set covering nine risk categories: \emph{safe}, \emph{destructive action}, \emph{sensitive data access}, \emph{external communication}, \emph{privilege misuse}, \emph{budget abuse}, \emph{ambiguous confirm}, \emph{ambiguous revise}, and \emph{multi-step escalation}. Each instance contains a natural-language request, an ordered structured plan, a binary safety label, and an intervention label from $\{\textsc{allow}, \textsc{block}, \textsc{confirm}, \textsc{revise}\}$.

\paragraph{IPI adversarial split.}
We construct $200$ paired indirect-prompt-injection cases across three templates (Section~\ref{sec:threat}): (A) injection in a retrieved RAG chunk ($40$ pairs), (B) injection in an earlier tool output ($40$ pairs), and (C) a plan-level evasion annotation claiming pre-approval ($20$ pairs). Each adversarial case has a matched benign control with the same topic, user, and tool category but without the unsafe step. This split tests whether \NEXUS correctly classifies the agent's resulting plan when the surrounding \emph{context} contains adversarial text.

The two synthetic splits evaluate different dimensions: structural detection (synthetic) and robustness to injection context (IPI). External benchmark generalization is evaluated on R-Judge~\cite{yuan2024r} (Section~\ref{sec:rjudge}) and AgentHarm~\cite{andriushchenko2025agentharm} (Section~\ref{sec:agentharm}); evaluation on organic agent traces is left for future work (Section~\ref{sec:limitations}).

\section{Main Results}
\label{sec:results}

Table~\ref{tab:cross-bench} summarizes the main results across three settings: the held-out synthetic split, R-Judge, and the indirect-prompt-injection (IPI) split. The synthetic split measures in-distribution structured-plan detection, while R-Judge serves as the main external validation benchmark. The IPI split is used as a robustness check: both \NEXUS and the rule-only monitor block all adversarial plans, showing that plan-level monitoring can catch injected behavior once it appears in the structured plan. AgentHarm is reported separately in \S\ref{sec:agentharm} because it tests a different boundary, where harmful and benign examples may share the same target tools.

On the synthetic split, \NEXUS achieves $F_1 = 0.949$ and $4$-class intervention accuracy of $0.641$. This matches rule-only detection in binary $F_1$ but substantially improves intervention routing over rule-only monitoring ($0.367$ to $0.641$). On R-Judge, \NEXUS reaches $F_1 = 0.861$ [$0.83, 0.89$], compared with $0.849$ for rule-only. The gain is concentrated in additional unsafe detections, especially in IoT trajectories after enabling the permanence-flag rule in \S\ref{sec:rjudge}. Per-category performance on R-Judge is strongest for Finance, Program, and Web ($F_1 \ge 0.89$), followed by Application ($0.852$), with IoT remaining the hardest category ($0.518$).

On AgentHarm ($n{=}352$), both \NEXUS and rule-only obtain $F_1 = 0.591$ [$0.53, 0.65$]. This occurs because paired harmful and benign examples often share the same target tools, so the structural plan does not contain the unsafe intent signal. We therefore treat AgentHarm as a threat-model boundary rather than as a regression.

\paragraph{Statistical significance.}
On the synthetic split, binary McNemar testing gives $p = 1.0$ because rule-only and \NEXUS make the same binary decisions. The scorer's effect appears in the $4$-class intervention task. A paired permutation test over intervention accuracy ($5{,}000$ permutations) gives $\Delta = +0.273$, $p < 0.001$: the scorer moves $35/128$ instances from \textsc{confirm} to \textsc{block}, and all $35$ have gold label \textsc{block}. On R-Judge, binary McNemar gives $\chi^2 = 3.125$, $p = 0.077$, corresponding to $7$ additional unsafe detections. Thus, \NEXUS differs most clearly from rule-only in intervention accuracy, per-category behavior, and R-Judge generalization, not in IPI or AgentHarm.

\begin{keyresult}{Scorer-gated demotion adds $+27.3$ pp}
\textbf{Scorer-gated demotion adds $\boldsymbol{+27.3}$ pp 4-class intervention accuracy} over rule-only monitoring ($0.367 \to 0.641$, permutation $p < 0.001$, $5{,}000$ permutations), while preserving identical binary $F_1 = 0.949$ and unsafe-caught rate.
\end{keyresult}

\begin{table}[t]
\centering
\scriptsize
\setlength{\tabcolsep}{2pt}
\renewcommand{\arraystretch}{1.12}

\begin{tabularx}{\linewidth}{lcccccc}
\toprule
\rowcolor{tabheader}
& \multicolumn{2}{c}{\textcolor{tabheadertext}{\textbf{Synthetic ($n{=}128$)}}}
& \multicolumn{2}{c}{\textcolor{tabheadertext}{\textbf{R-Judge ($n{=}564$)}}}
& \multicolumn{2}{c}{\textcolor{tabheadertext}{\textbf{IPI ($n{=}200$)}}} \\
\cmidrule(lr){2-3} \cmidrule(lr){4-5} \cmidrule(l){6-7}
\rowcolor{tabheader}
\textcolor{tabheadertext}{\textbf{Monitor}} &
\textcolor{tabheadertext}{\textbf{$F_1$}} &
\textcolor{tabheadertext}{\textbf{Int.-Acc}} &
\textcolor{tabheadertext}{\textbf{$F_1$}} &
\textcolor{tabheadertext}{\textbf{Prec.}} &
\textcolor{tabheadertext}{\textbf{$F_1$}} &
\textcolor{tabheadertext}{\textbf{Int.-Acc}} \\
\midrule
\rowcolor{tabalt}
Rule-only    & 0.949 & 0.367 & 0.849 & 0.858 & 1.000 & 1.000 \\
Learned-only & 0.549 & 0.195 & --    & --    & 1.000 & 1.000 \\
\rowcolor{tabnexus}
\textbf{\NEXUS}
& \makecell{\textbf{0.949}\\{\tiny [.92,.98]}}
& \makecell{\textbf{0.641}\\{\tiny [.56,.72]}}
& \makecell{\textbf{0.861}\\{\tiny [.83,.89]}}
& \makecell{\textbf{0.858}\\{\tiny [.82,.90]}}
& \makecell{0.995\\{\tiny [.98,1.00]}}
& \textbf{0.995} \\
\bottomrule
\end{tabularx}

\caption{Headline results across synthetic, R-Judge, and IPI settings. Bracketed values are $95\%$ bootstrap CIs for \NEXUS. Synthetic and R-Judge measure in-distribution and external performance, respectively; IPI is a robustness check for plan-level monitoring under injected context.}
\label{tab:cross-bench}
\end{table}


\section{Extended Empirical Analysis}
\label{sec:extended}

\subsection{Category-Level Behavior}
\label{sec:per-cat}

Table~\ref{tab:per-cat-app} reports binary $F_1$ by risk category on the synthetic test split. \NEXUS shows its largest gains when unsafe behavior appears inside tool arguments or plan structure. Compared with rule-only monitoring, \NEXUS improves \emph{sensitive data access} from $F_1 = 0.273$ to $1.000$, because argument inspection detects credentials and sensitive paths that category-level rules miss. Compared with learned-only monitoring, \NEXUS improves \emph{budget abuse}, \emph{ambiguous confirm}, and \emph{ambiguous revise} by $F_1 = +1.000$ each. This shows that a scalar risk score alone does not reliably capture structural signals such as exceeded budgets or under-specified intent. The main remaining weakness is \emph{multi-step escalation}, where single-turn evaluation reaches only $F_1 = 0.500$.

\subsection{Session-Aware Multi-Turn Monitoring}
\label{sec:multiturn}

The low score on \emph{multi-step escalation} comes from evaluating each turn independently. To test whether session context resolves this failure mode, we evaluate \NEXUS on a session-aware benchmark of $120$ ordered sessions: cross-step exfiltration ($35$), escalating privilege ($20$), incremental sensitive access ($20$), repeated unsafe intent ($20$), and legitimate multi-turn controls ($25$).

We extend the \textsc{confirm} branch of $\Pi$ with four session-state features: an exposed-entities set, a cumulative permission set, a repeated-unsafe-attempt counter, and a cumulative-cost accumulator. No additional learned component is added. Table~\ref{tab:multiturn} shows that session memory raises critical-turn detection from $0/95$ to $95/95$ across the unsafe sessions, with zero false positives on the $25$ legitimate controls. This result suggests that the multi-step failure is mainly due to missing session state, not to the rule vocabulary alone.

\begin{keyresult}{Multi-turn gap: $0/95 \to 95/95$}
\textbf{Session memory closes the multi-turn detection gap entirely: $\boldsymbol{0/95 \to 95/95}$ critical turns caught, $\boldsymbol{0}$ false positives on $25$ legitimate controls.} No additional learned component was required; lifting four lightweight session-state features into $\Pi$'s \textsc{confirm} branch is sufficient.
\end{keyresult}

\begin{table}[t]
\centering
\small
\setlength{\tabcolsep}{4pt}
\renewcommand{\arraystretch}{1.08}
\begin{tabularx}{\linewidth}{Xcc}
\toprule
\rowcolor{tabheader}
\textcolor{tabheadertext}{\textbf{Metric}} &
\textcolor{tabheadertext}{\textbf{No session memory}} &
\textcolor{tabheadertext}{\textbf{With session memory}} \\
\midrule
\rowcolor{tabalt}
TP / FN (critical turn) & \textcolor{nored}{0 / 95} & \textcolor{okgreen}{\textbf{95 / 0}} \\
FP / TN (legitimate ctrl) & 0 / 25 & \textcolor{okgreen}{\textbf{0 / 25}} \\
\midrule
\rowcolor{tabalt}
Precision / Recall / $F_1$ & \textcolor{nored}{0.00 / 0.00 / 0.00} & \textcolor{okgreen}{\textbf{1.00 / 1.00 / 1.00}}$^{\dagger}$ \\
Control allow rate & 1.000 & 1.000$^{\ddagger}$ \\
\bottomrule
\end{tabularx}
\caption{Session-aware multi-turn evaluation on $120$ authored sessions. Session memory catches all $95$ unsafe critical turns with zero false positives on $25$ legitimate controls. $^{\dagger}$Wilson $95\%$ CI for $95/95$: $[0.961, 1.000]$. $^{\ddagger}$Wilson $95\%$ CI for $25/25$: $[0.864, 1.000]$.}
\label{tab:multiturn}
\end{table}

\subsection{External Generalization on R-Judge}
\label{sec:rjudge}

\paragraph{Benchmark setup.}
We evaluate external generalization on R-Judge~\cite{yuan2024r}, a public multi-turn agent-safety benchmark with $564$ usable records across Application, Finance, IoT, Program, and Web, after excluding seven leaked records. Each record provides an agent--environment trajectory, binary safety label, and attack type. We parse structured \texttt{agent.action} strings into \NEXUS \texttt{PlanStep} objects; $491/571$ records yield real structured plans, while $80$ dialogue-only Application records are mapped to placeholder steps. Tool metadata is inferred using keyword tables.

\begin{table}[t]
\centering
\scriptsize
\setlength{\tabcolsep}{2.5pt}
\renewcommand{\arraystretch}{1.15}
\begin{tabular}{p{0.23\columnwidth}p{0.19\columnwidth}p{0.20\columnwidth}p{0.17\columnwidth}p{0.13\columnwidth}}
\toprule
\rowcolor{tabheader}
\textcolor{tabheadertext}{\textbf{Subset}} &
\textcolor{tabheadertext}{\textbf{Monitor}} &
\textcolor{tabheadertext}{\textbf{$F_1$ [95\% CI]}} &
\textcolor{tabheadertext}{\textbf{Prec.}} &
\textcolor{tabheadertext}{\textbf{Rec.}} \\
\midrule
\rowcolor{tabalt}
Full ($n{=}564$) & Rule-only & 0.849 & 0.858 & 0.841 \\
\rowcolor{tabnexus}
Full ($n{=}564$) & \NEXUS & \textbf{0.861}\\[-1pt] {\tiny [0.831, 0.888]} & 0.858\\[-1pt] {\tiny [0.821, 0.895]} & 0.864\\[-1pt] {\tiny [0.823, 0.901]} \\
Real-plan ($n{=}491$) & \NEXUS & 0.832 & 0.831 & 0.834 \\
\bottomrule
\end{tabular}
\caption{External generalization on R-Judge. \NEXUS improves over the rule-only monitor on the full benchmark, increasing recall while preserving precision. The real-plan subset includes only records that could be parsed into structured tool plans, excluding dialogue-only placeholder cases.}
\label{tab:rjudge}
\end{table}

\paragraph{Results.}
Table~\ref{tab:rjudge} shows that \NEXUS reaches $F_1 = 0.861$, improving over rule-only by $+0.012$ while remaining below the synthetic split result ($0.949$), as expected under distribution shift. The gain is mainly from seven additional unsafe detections (McNemar $\chi^2 = 3.125$, $p = 0.077$). Retraining the scorer on R-Judge with $5$-fold CV raises $F_1$ to $0.892$ versus $0.848$ for the synthetic-trained baseline, indicating that the feature representation captures transferable safety signal.

\paragraph{Category analysis.}
\NEXUS performs best on Finance, Program, and Web, where risks appear as parseable transfer, delete, transmit, or execution actions. IoT remains hardest: the permanence-flag rule improves IoT $F_1$ from $0.462$ to $0.518$ and overall R-Judge $F_1$ from $0.848$ to $0.861$, but remaining errors involve cross-step identity matching and incomplete smart-home semantics. Thus, R-Judge supports external validity for structured tool-call monitoring while highlighting the need for richer domain-specific semantics.


\subsection{Threat-Model Boundary on AgentHarm}
\label{sec:agentharm}

AgentHarm~\cite{andriushchenko2025agentharm} contains $176$ harmful and $176$ benign tool-use behaviors across eight categories. Each example includes a free-text prompt and a list of target functions. We map each target function to a \NEXUS \texttt{PlanStep} using the same keyword-based classifier used for R-Judge, but we do not include the prompt text in the plan.

On this $352$-instance test set, \NEXUS obtains $F_1 = 0.591$ [$0.53, 0.65$], with precision $0.508$ and recall $0.705$. Rule-only monitoring obtains the same $F_1$. This result is expected because AgentHarm often pairs harmful and benign examples that use the same target tools. In such cases, the unsafe signal is mainly expressed in the prompt intent, not in the structural tool sequence.

\paragraph{Scope analysis.}
AgentHarm therefore tests a boundary of plan-level monitoring. \NEXUS is designed to detect unsafe structure in plans, arguments, permissions, and side effects; it is not designed to infer latent harmful intent from the prompt alone. When harmfulness depends on prompt-level intent while the tool plan remains structurally similar, prompt-aware or argument-aware methods such as ToolSafe~\cite{mou2026toolsafe} are the more appropriate comparison class. The identical performance of \NEXUS and rule-only monitoring further supports our design claim: the learned risk score is used for borderline intervention routing, not for latent-intent detection.

\begin{takeaway}{AgentHarm parity: a structural boundary}
When paired harmful and benign behaviours share the same structural tool plan, the unsafe signal lives in prompt-level intent, not in the plan; plan-level monitors of any kind are structurally blind here. \NEXUS is positioned as one layer of a defence stack, complementary to prompt-aware safeguards.
\end{takeaway}

\subsection{Runtime Efficiency}
\label{sec:runtime}

Table~\ref{tab:latency} reports the runtime cost of \NEXUS on the $128$-instance synthetic test split. We repeat the evaluation three times, yielding $384$ total decisions on a single CPU core without batching. \NEXUS is lightweight: it achieves a median latency of $0.205$\,ms and a P99 latency of $0.456$\,ms, corresponding to approximately $4{,}349$ decisions per second. This overhead is negligible compared with a typical LLM agent turn, which often takes hundreds or thousands of milliseconds. Therefore, \NEXUS can be inserted before tool execution without requiring a GPU or noticeably slowing the agent loop.

\begin{keyresult}{$0.205$\,ms median latency, no GPU}
Under $0.1\%$ overhead on a typical agent loop; dominant cost is \texttt{predict\_proba} on a $9$-D logistic regression.
\end{keyresult}

\begin{table}[t]
\centering
\small
\begin{tabular}{lc}
\toprule
\rowcolor{tabheader}
\textcolor{tabheadertext}{\textbf{Metric}} & \textcolor{tabheadertext}{\textbf{Value}} \\
\midrule
\rowcolor{tabalt}
Median latency & \textbf{$0.205$\,ms} \\
P90 latency & $0.324$\,ms \\
\rowcolor{tabalt}
P95 latency & $0.362$\,ms \\
P99 latency & $0.456$\,ms \\
\rowcolor{tabnexus}
Throughput & \textbf{${\sim}4{,}349$ decisions/sec} \\
\bottomrule
\end{tabular}
\caption{Runtime latency of \NEXUS on the synthetic test split. Results are computed over $384$ evaluations from three repeated runs on a single CPU core without batching.}
\label{tab:latency}
\end{table}

\subsection{Comparison with LLM Judges}
\label{sec:llm-judge}

We compare \NEXUS with GPT-4o on the same $200$-instance \NEXUS-Stress benchmark, treating GPT-4o as a strong LLM-judge baseline. As shown in Table~\ref{tab:llm-judge}, GPT-4o obtains higher binary detection and intervention accuracy, but at substantially higher latency. \NEXUS reaches $92\%$ of GPT-4o's binary $F_1$ and $78\%$ of its intervention accuracy while running approximately $4{,}800{\times}$ faster. GPT-4o also struggles with fine-grained intervention routing, with per-class $F_1$ of $0.455$ for \textsc{confirm} and $0.405$ for \textsc{revise}. This suggests that the hardest part of \NEXUS-Stress is not only unsafe-plan detection, but deciding the correct intervention level.

\begin{table}[t]
\centering
\small
\setlength{\tabcolsep}{4pt}
\begin{tabular}{lccc}
\toprule
\rowcolor{tabheader}
\textcolor{tabheadertext}{\textbf{Monitor}} &
\textcolor{tabheadertext}{\textbf{Binary $F_1$}} &
\textcolor{tabheadertext}{\textbf{Int.-Acc}} &
\textcolor{tabheadertext}{\textbf{Latency}} \\
\midrule
\rowcolor{tabalt}
GPT-4o & 0.956 & 0.555 & \textcolor{nored}{${\sim}1{,}200$\,ms} \\
\rowcolor{tabnexus}
\textbf{\NEXUS} & \textbf{0.881} & \textbf{0.435} & \textcolor{okgreen}{\textbf{0.205\,ms}} \\
\midrule
\rowcolor{taboos}
NEXUS / GPT-4o & $92\%$ & $78\%$ & \textcolor{okgreen}{\textbf{${\sim}4{,}800{\times}$ faster}}$^{\dagger}$ \\
\bottomrule
\end{tabular}
\caption{Comparison with GPT-4o on the $200$-instance \NEXUS-Stress benchmark. \NEXUS is less accurate than GPT-4o but is several orders of magnitude faster, making it more suitable for runtime monitoring before tool execution. $^{\dagger}$Speed ratio uses a conservative $0.25$\,ms \NEXUS latency; the measured median latency is $0.205$\,ms.}
\label{tab:llm-judge}
\end{table}

\subsection{Adversarial Stress Testing}
\label{sec:stress}

We evaluate \NEXUS on \textbf{\NEXUS-Stress}, a $200$-example adversarial benchmark designed without matching the rule-set vocabulary. The benchmark covers $20$ scenario categories across $11$ domains; Appendix~\ref{app:stress} provides the construction and audit details.

\NEXUS achieves binary $F_1 = 0.881$, with precision $0.829$ and recall $0.939$. Table~\ref{tab:stress-cm} provides the $4$-class confusion matrix, which gives a more detailed view of intervention behavior. Among the $94$ gold \textsc{confirm} and \textsc{revise} examples, \NEXUS frequently predicts \textsc{confirm} ($96$ predictions vs.\ $54$ gold). This shows a conservative pattern: when the system is uncertain, it tends to ask for user confirmation instead of blocking automatically.

The main limitation is that \NEXUS does not predict \textsc{revise} on rule-blind inputs. This exposes a middle-severity coverage gap, where the plan is not safe enough to allow but does not clearly match existing revision-triggering rules. We discuss this limitation in \S\ref{sec:limitations}, with per-category details in Appendix~\ref{app:stress}.

\begin{table}[t]
\centering
\scriptsize
\setlength{\tabcolsep}{4pt}
\begin{tabular}{lcccc}
\toprule
\rowcolor{tabheader}
\textcolor{tabheadertext}{\textbf{gold $\downarrow$ / pred $\rightarrow$}} &
\textcolor{tabheadertext}{\textbf{ALLOW}} &
\textcolor{tabheadertext}{\textbf{BLOCK}} &
\textcolor{tabheadertext}{\textbf{CONFIRM}} &
\textcolor{tabheadertext}{\textbf{REVISE}} \\
\midrule
\rowcolor{tabalt}
ALLOW   ($n=35$)  & \cellcolor{tabnexus}\textbf{19} & 3  & 13 & 0 \\
BLOCK   ($n=71$)  & 6  & \cellcolor{tabnexus}\textbf{30} & 35 & 0 \\
\rowcolor{tabalt}
CONFIRM ($n=54$)  & 2  & 4  & \cellcolor{tabnexus}\textbf{48} & 0 \\
REVISE  ($n=40$)  & 15 & 0  & 25 & \cellcolor{taboos}\textcolor{nored}{\textbf{0}} \\
\bottomrule
\end{tabular}
\caption{\NEXUS-Stress $4$-class confusion matrix. \NEXUS shows conservative behavior on ambiguous cases by often selecting \textsc{confirm}, but it does not predict \textsc{revise}, indicating a middle-severity coverage gap.}
\label{tab:stress-cm}
\end{table}

\subsection{Additional Results in the Appendix}
\label{sec:extended-results}

The appendix reports supporting analyses that complement the main results: confusion matrices (Appendix~\ref{app:confusion}), threshold and safety--utility sweeps (Appendix~\ref{app:thresholds}), strengthened learned-only baselines (Appendix~\ref{app:strong-learned}), policy ablations (Appendix~\ref{app:policy-ablation}), IPI v2 across five injection styles (Appendix~\ref{app:ipi-v2}), scorer-based reclassification analysis (Appendix~\ref{app:reclass}), error analysis (Appendix~\ref{app:errors}), and the full \NEXUS-Stress breakdown (Appendix~\ref{app:stress}). The key findings from these analyses are summarized in Tables~\ref{tab:cross-bench},~\ref{tab:rjudge}, and~\ref{tab:stress-cm}.

\section{Discussion}
\label{sec:discussion}
\label{sec:gen-structured}

Runtime safety for tool-using agents is best viewed as a plan-level control problem. Once agents invoke tools, safety-relevant signals appear in tool choices, arguments, permissions, side effects, and execution order. \NEXUS uses this structured plan representation to inspect executable behavior before tool calls are committed and to ground interventions in explicit rules, argument findings, or calibrated risk scores.

The results show that the hybrid design is useful. Rules enforce hard constraints, argument inspection detects sensitive values inside tool inputs, and the calibrated scorer improves intervention routing. In particular, the scorer mainly helps distinguish \textsc{block} from \textsc{confirm}, rather than serving as the primary unsafe-plan detector.

The main challenge is fine-grained intervention selection. Binary unsafe detection is easier than deciding whether to \textsc{block}, \textsc{confirm}, or \textsc{revise}. The \NEXUS-Stress results show that medium-severity and rule-blind cases remain difficult, especially when revision is more appropriate than confirmation.

The R-Judge and AgentHarm results clarify the boundary of plan-level monitoring. \NEXUS works best when risk is visible in executable structure, such as deletion, transmission, sensitive access, or permission misuse. It is weaker when harmfulness depends mainly on latent prompt intent. Thus, \NEXUS should complement prompt-aware and intent-aware safeguards rather than replace them.

Finally, the sub-millisecond CPU latency makes \NEXUS practical as a runtime governance layer before tool execution. Its modular design allows rules, argument checkers, and thresholds to be updated without retraining the full system.

\section{Conclusion}
\label{sec:conclusion}

We presented \NEXUS, a runtime safety monitor for tool-using LLM agents that combines nine deterministic rules, argument-level inspection, and a calibrated risk scorer under a four-class scorer-gated demotion policy $\Pi$. The scorer is used for intervention selection rather than binary detection: while rules provide perfect recall, the scorer improves $4$-class intervention accuracy by $+27.3$ pp by distinguishing BLOCK from CONFIRM when a critical rule fires. Across benchmarks, \NEXUS achieves strong safety performance with practical runtime overhead. On the $128$-instance synthetic split, it obtains $F_1 = 0.949$ and $4$-class intervention accuracy of $0.641$. On R-Judge ($n{=}564$), it reaches $F_1 = 0.861$ [$0.83, 0.89$], improving over rule-only monitoring, with the permanence-flag rule raising IoT performance from $0.462$ to $0.518$. On AgentHarm, \NEXUS matches rule-only performance ($F_1 = 0.591$), highlighting a deliberate threat-model boundary when harmful and benign behaviours share target tools. On IPI, both monitors achieve $0\%$ ASR at $99\%$ control allow, showing robustness to injection text at the structured-plan level. On \NEXUS-Stress, \NEXUS reaches $F_1 = 0.881$ and remains competitive with GPT-4o while requiring only $0.205$\,ms median latency and zero marginal cost. Future work should expand medium-severity rule coverage to improve CONFIRM/REVISE routing and validate session-aware multi-turn monitoring on organic agent traces. We release the code, benchmarks, retrained $9$-D risk scorer, and CI pipelines.

\section*{Limitations and Ethical Considerations}
\label{sec:limitations}

\paragraph{Scope.}
Several evaluations use author-generated templates, so in-distribution results should be viewed as upper bounds rather than deployment estimates. R-Judge ($F_1 = 0.861$) and \NEXUS-Stress ($F_1 = 0.881$) provide stronger out-of-distribution evidence, but live-trace evaluation is still needed. \NEXUS currently supports seven tools; broader registries require new side-effect and permission encodings. Human evaluation of explanations/revisions is also left for future work.

\paragraph{Coverage gap.}
\NEXUS-Stress shows that rule-blind medium-severity cases are difficult: $\Pi$ predicts no REVISE on $40$ gold REVISE examples and defaults to CONFIRM. Future rules should target scope-tightening and disambiguation patterns.

\paragraph{Intervention difficulty.}
CONFIRM/REVISE routing remains challenging, with even GPT-4o reaching only $0.555$ intervention accuracy on adversarial data. Hybrid fast-filter and LLM-judge designs may help.

\paragraph{Ethics.}
The benchmarks contain no real PII and use fabricated credentials/URLs. We release templates following prior benchmark norms, as exposing failure modes primarily supports defensive research.

\bibliography{custom}

\newpage

\appendix
\onecolumn
\section*{Appendix}

This appendix provides supplementary experimental results, per-category analyses, method details, deployment notes, and reproducibility, ethics, and AI-usage statements. The material is organized thematically to help readers locate supporting evidence for the main paper.

\paragraph{Algorithm and method details.}
\begin{itemize}[leftmargin=*,itemsep=2pt,topsep=2pt]
\item \textbf{Appendix~\ref{app:algorithm}} presents the complete runtime monitoring algorithm with scorer-gated demotion.
\item \textbf{Appendix~\ref{app:features}} defines the $9$-dimensional plan-level feature vector $\phi(P)$.
\item \textbf{Appendix~\ref{app:calibration-fig}} reports calibration metrics and the reliability diagram.
\end{itemize}

\paragraph{Extended experimental results.}
\begin{itemize}[leftmargin=*,itemsep=2pt,topsep=2pt]
\item \textbf{Appendix~\ref{app:confusion}} reports binary and 4-class confusion matrices for \NEXUS on the synthetic test split.
\item \textbf{Appendix~\ref{app:thresholds}} reports the $(\tau_b, \tau_c)$ threshold sweep and the $5{\times}5$ safety--utility sensitivity analysis.
\item \textbf{Appendix~\ref{app:strong-learned}} reports strengthened learned-only baselines.
\item \textbf{Appendix~\ref{app:policy-ablation}} reports policy ablations with bootstrap confidence intervals.
\item \textbf{Appendix~\ref{app:ipi-v2}} reports IPI robustness across five injection styles.
\item \textbf{Appendix~\ref{app:multiturn-examples}} provides qualitative multi-turn examples and per-category critical-turn detection.
\item \textbf{Appendix~\ref{app:reclass}} analyzes reclassifications produced by the learned scorer.
\item \textbf{Appendix~\ref{app:errors}} summarizes the main error categories.
\end{itemize}

\paragraph{Per-category and benchmark breakdowns.}
\begin{itemize}[leftmargin=*,itemsep=2pt,topsep=2pt]
\item \textbf{Appendix~\ref{app:per-cat-table}} reports per-category synthetic test performance.
\item \textbf{Appendix~\ref{app:rjudge-category}} reports the per-category R-Judge breakdown.
\item \textbf{Appendix~\ref{app:rjudge-iot}} provides trace-level analysis of R-Judge IoT failures.
\item \textbf{Appendix~\ref{app:agentharm-table}} reports per-category AgentHarm results.
\item \textbf{Appendix~\ref{app:stress}} reports the full \NEXUS-Stress audit, metrics, per-category results, difficulty breakdown, and failure modes.
\end{itemize}

\paragraph{Deployment, implementation, and environment.}
\begin{itemize}[leftmargin=*,itemsep=2pt,topsep=2pt]
\item \textbf{Appendix~\ref{app:deployment-posture}} discusses deployment posture and future extensions.
\item \textbf{Appendix~\ref{app:implementation}} documents operational inputs, decision outputs, controlled tool categories, and determinism settings.
\item \textbf{Appendix~\ref{app:prompts}} describes the prompt templates and links to the released code.
\item \textbf{Appendix~\ref{app:env}} documents software, hardware, and storage requirements.
\end{itemize}

\paragraph{Reproducibility, ethics, and AI usage.}
\begin{itemize}[leftmargin=*,itemsep=2pt,topsep=2pt]
\item \textbf{Appendix~\ref{app:reproducibility}} provides the reproducibility statement, including released artifacts, seeds, splits, and commands.
\item \textbf{Appendix~\ref{app:ethics}} provides the ethics statement.
\item \textbf{Appendix~\ref{app:ai-usage}} provides the AI-usage statement.
\end{itemize}

\newpage

\section{Runtime Monitoring Algorithm}
\label{app:algorithm}

Algorithm~\ref{alg:nexus} gives the complete runtime procedure used by \NEXUS. The algorithm is designed to keep the three safety signals logically separate before combining them through the intervention policy. First, deterministic rules identify explicit policy violations in the structured plan. Second, the argument inspector detects risky values inside tool inputs, such as credentials, sensitive paths, or suspicious URLs. Third, the calibrated risk scorer estimates the plan-level risk $\rho$. These signals are then merged into the violation set $V$ and resolved through a top-down intervention cascade.

The ordering of the cascade is intentional. Multiple critical violations always produce a hard block, because they indicate high-confidence unsafe behavior. When exactly one critical violation is present, the calibrated risk score determines whether the plan should be blocked or routed to user confirmation. This scorer-gated demotion step prevents every single critical violation from becoming an automatic block while still allowing high-risk cases to be stopped. High-severity violations are routed to confirmation, medium-severity violations are routed to revision, and the learned scorer provides a residual confirmation signal when no rule or argument violation is present.

\begin{coreidea}{Cascade order: rules first, scorer second}
The cascade is non-commutative by design. The $n_{\mathrm{crit}} \ge 2$ rule precedes scorer-gated demotion so that the scorer can never override an unambiguous signal. The scorer's authority is bounded: it only adjudicates between $\BLOCK$ and $\CONFIRM$ when exactly one critical violation fires, and only contributes a residual $\CONFIRM$ when no rule violation is present. This is what makes hybrid monitoring auditable.
\end{coreidea}

\begin{algorithm}[t]
\caption{\NEXUS runtime monitoring with scorer-gated demotion}
\label{alg:nexus}
\begin{algobox}
\begin{algorithmic}[1]
\Require plan $P$, calibrated risk scorer $f$, deterministic rule set $\mathcal{R}$,
         argument inspector $\mathcal{A}$, thresholds $(\tau_b, \tau_c)$
\Ensure intervention $i \in \{\ALLOW, \BLOCK, \CONFIRM, \REVISE\}$
        and justification $J$

\State $V_R \gets \mathcal{R}(P)$ \Comment{\textcolor{catmot}{rule-based violations}}
\State $V_A \gets \mathcal{A}(P)$ \Comment{\textcolor{catmot}{argument-level violations}}
\State $V \gets V_R \cup V_A$
\State $\rho \gets f.\textsc{Predict}(P)$ \Comment{\textcolor{catmot}{calibrated plan risk}}
\State $n_c \gets |\{v \in V : \mathrm{sev}(v) = \textsc{crit}\}|$

\If{$n_c \ge 2$}
    \State \Return $(\BLOCK,\ \textsc{Justify}(V_{\text{crit}}, \rho))$
\EndIf

\If{$n_c = 1 \wedge \rho \ge \tau_b$}
    \State \Return $(\BLOCK,\ \textsc{Justify}(V_{\text{crit}}, \rho))$
\EndIf

\If{$n_c = 1 \wedge \rho < \tau_b$}
    \State \Return $(\CONFIRM,\ \textsc{Justify}(V_{\text{crit}}, \rho))$
\EndIf

\If{$\exists\, v \in V : \mathrm{sev}(v) = \textsc{high}$}
    \State \Return $(\CONFIRM,\ \textsc{Justify}(V_{\text{high}}, \rho))$
\EndIf

\If{$\exists\, v \in V : \mathrm{sev}(v) = \textsc{med}$}
    \State \Return $(\REVISE,\ \textsc{Justify}(V_{\text{med}}, \rho))$
\EndIf

\If{$\rho \ge \tau_c$}
    \State \Return $(\CONFIRM,\ \textsc{Justify}(\emptyset, \rho))$
\EndIf

\State \Return $(\ALLOW,\ \textsc{Justify}(\emptyset, \rho))$
\end{algorithmic}
\end{algobox}
\end{algorithm}

\section{Feature Vector Specification}
\label{app:features}

\paragraph{Feature vector $\phi(P)$.}
We train a logistic-regression risk scorer on a 9-dimensional plan-level feature vector. Each plan $P$ is mapped to $\phi(P) \in \mathbb{R}^{9}$:
\begin{enumerate}[leftmargin=*,itemsep=0pt,topsep=2pt]
\item $\phi_1$: plan length $|P|$ (raw integer step count).
\item $\phi_2$: indicator $\mathbf{1}[\exists\, s_i \in P : s_i.\textit{irreversible}]$.
\item $\phi_3$: number of file-category steps.
\item $\phi_4$: number of database-category steps.
\item $\phi_5$: number of network-category steps.
\item $\phi_6$: indicator $\mathbf{1}[\exists\, s_i : s_i.\textit{sensitive}]$.
\item $\phi_7$: total estimated cost $\sum_i s_i.\textit{cost}$.
\item $\phi_8$: tool-category diversity $|\{s_i.\textit{category}\}|$.
\item $\phi_9$: indicator $\mathbf{1}[\exists\, s_i : \textsc{delete} \in s_i.\textit{side\_effects}]$.
\end{enumerate}
All features are z-score normalised using the training-set mean and standard deviation (a single \texttt{StandardScaler} fit on the training split is persisted with the model). An earlier $10$-dimensional version included a normalised-plan-length feature that was numerically identical to $\phi_1$ on the deployed benchmark; it was dropped following peer-reviewer feedback, and the model retrained on the resulting $9$-D vector.

\section{Risk-Score Calibration}
\label{app:calibration-fig}

\begin{figure}[t]
\centering
\includegraphics[width=0.7\textwidth]{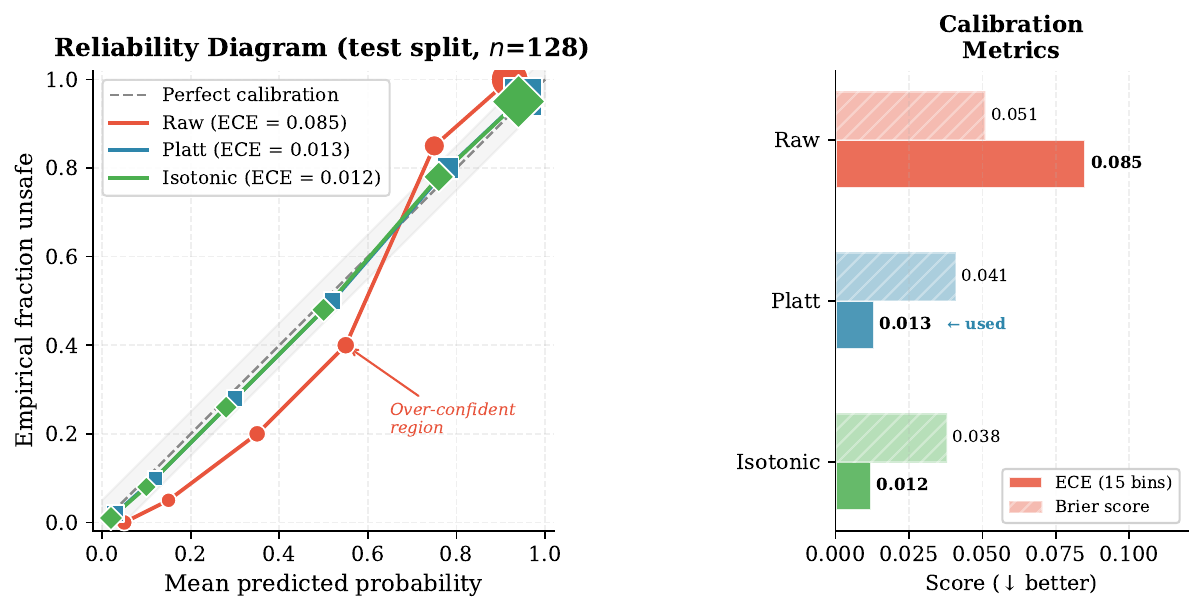}
\caption{Reliability diagram for the learned risk scorer on the $128$-instance held-out test split. The diagonal line indicates perfect calibration, and marker size reflects the number of examples in each bin. The raw scorer is over-confident in the mid-probability range, while post-hoc calibration moves the predicted probabilities closer to the empirical unsafe rates.}
\label{fig:calibration}
\end{figure}

Table~\ref{tab:calibration} reports calibration metrics for the learned risk scorer on the held-out test split, and Figure~\ref{fig:calibration} shows the corresponding reliability diagram. Both post-hoc calibration methods substantially reduce expected calibration error (ECE) compared with the raw logistic-regression probabilities. We use Platt scaling as the default because it provides strong calibration while being more stable on the small calibration split. Although isotonic regression obtains a slightly lower Brier score, Platt scaling offers the best ECE among the reported settings and avoids overfitting risk on limited calibration data.

\begin{keyresult}{Platt scaling: $\approx 6.5\times$ ECE reduction}
ECE drops from $0.085$ (raw logistic regression) to $\mathbf{0.013}$ after Platt scaling, so the thresholds $(\tau_b, \tau_c)$ operate on a probability scale that meaningfully reflects empirical unsafe rates. This is the foundation that makes the scorer-gated demotion step in $\Pi$ behaviour-correct rather than just statistically convenient.
\end{keyresult}

\begin{table}[t]
\centering
\small
\begin{tabular}{lcc}
\toprule
\rowcolor{tabheader}
\textcolor{tabheadertext}{\textbf{Calibration}} &
\textcolor{tabheadertext}{\textbf{ECE (15 bins) $\downarrow$}} &
\textcolor{tabheadertext}{\textbf{Brier $\downarrow$}} \\
\midrule
\rowcolor{tabalt}
Raw (uncalibrated)  & \textcolor{nored}{0.085} & 0.051 \\
\rowcolor{tabnexus}
\textbf{Platt scaling} (used) & \textcolor{okgreen}{\textbf{0.013}} & 0.041 \\
Isotonic regression & 0.012 & \textcolor{okgreen}{\textbf{0.038}} \\
\bottomrule
\end{tabular}
\caption{Calibration quality of the learned risk scorer on the $128$-instance held-out test split. Post-hoc calibration substantially improves probability reliability compared with the raw logistic-regression scorer. Platt scaling is used in \NEXUS because it achieves the lowest ECE among the reported settings and is less prone to overfitting on the small calibration split, while isotonic regression obtains the lowest Brier score.}
\label{tab:calibration}
\end{table}

\paragraph{Reliability-diagram caveat.}

Figure~\ref{fig:calibration} visualizes the calibration behavior of the raw and post-hoc calibrated risk scores. The raw scorer deviates from the ideal diagonal in the mid-probability range, indicating over-confidence. After calibration, the predicted probabilities move closer to the empirical unsafe rates. Because the diagram is computed on only $128$ held-out examples with $15$ equal-width bins, some bins contain few samples; therefore, we treat the plot as a qualitative diagnostic and rely on ECE and Brier score in Table~\ref{tab:calibration} as the primary calibration summaries.

\section{Confusion Matrices}
\label{app:confusion}

\begin{figure}[t]
\centering
\includegraphics[width=0.7\textwidth]{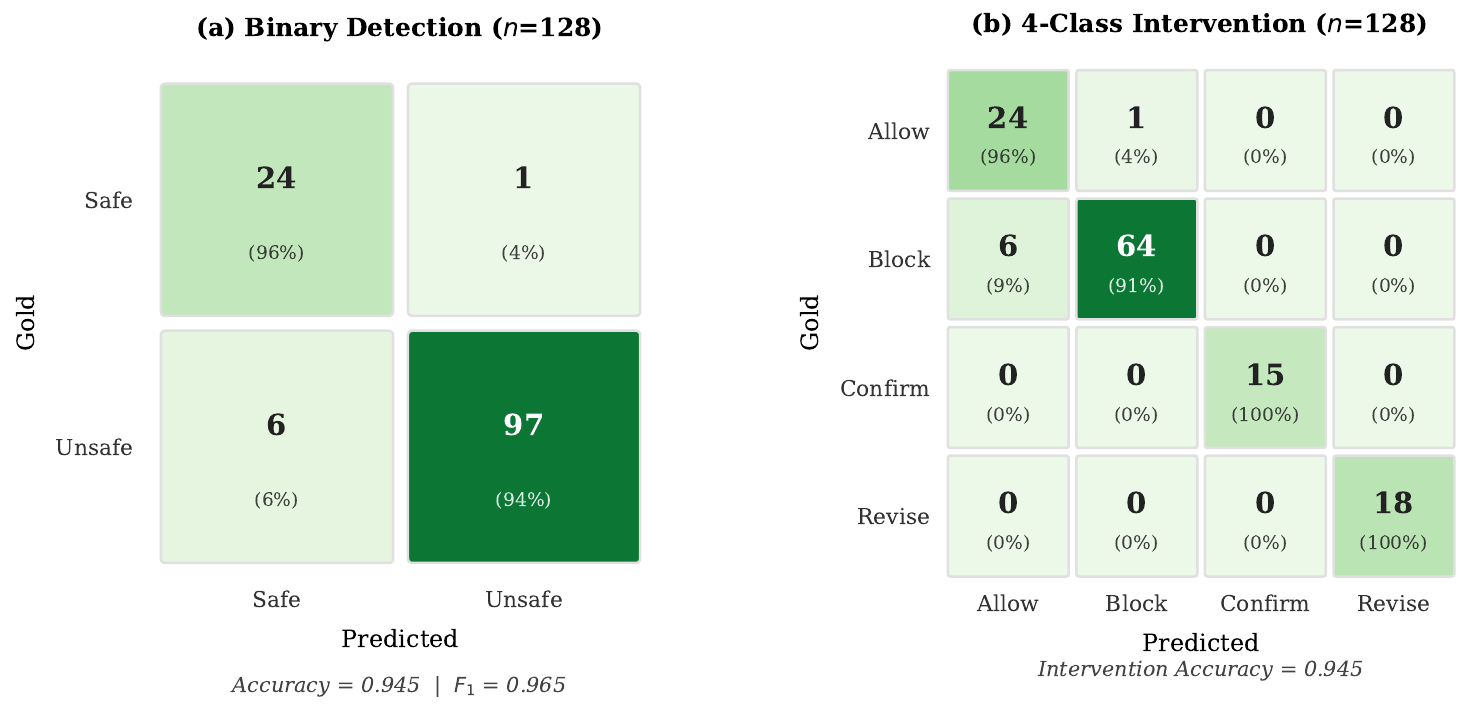}
\caption{Confusion matrices for \NEXUS on the $128$-instance synthetic test split. The binary matrix shows safe/unsafe detection performance, while the 4-class matrix shows intervention routing across \textsc{allow}, \textsc{block}, \textsc{confirm}, and \textsc{revise}. Errors are concentrated in gold \textsc{block} cases predicted as \textsc{allow}.}
\label{fig:cm}
\end{figure}

Figure~\ref{fig:cm} reports the binary and 4-class confusion matrices for \NEXUS on the $128$-instance synthetic test split. In the binary setting, \NEXUS correctly identifies $97$ of $103$ unsafe plans and allows $24$ of $25$ safe plans. The remaining errors consist of one safe plan flagged as unsafe and six unsafe plans predicted as safe. In the 4-class setting, \NEXUS correctly routes all \textsc{confirm} and \textsc{revise} cases, while most residual error is concentrated in the \textsc{block} class, where six gold \textsc{block} plans are predicted as \textsc{allow}. This pattern shows that the main failure mode is not confusion between confirmation and revision, but missed high-severity cases with weak structural signals.

\section{Threshold and Safety--Utility Sensitivity}
\label{app:thresholds}

\begin{figure}[t]
\centering
\includegraphics[width=0.7\textwidth]{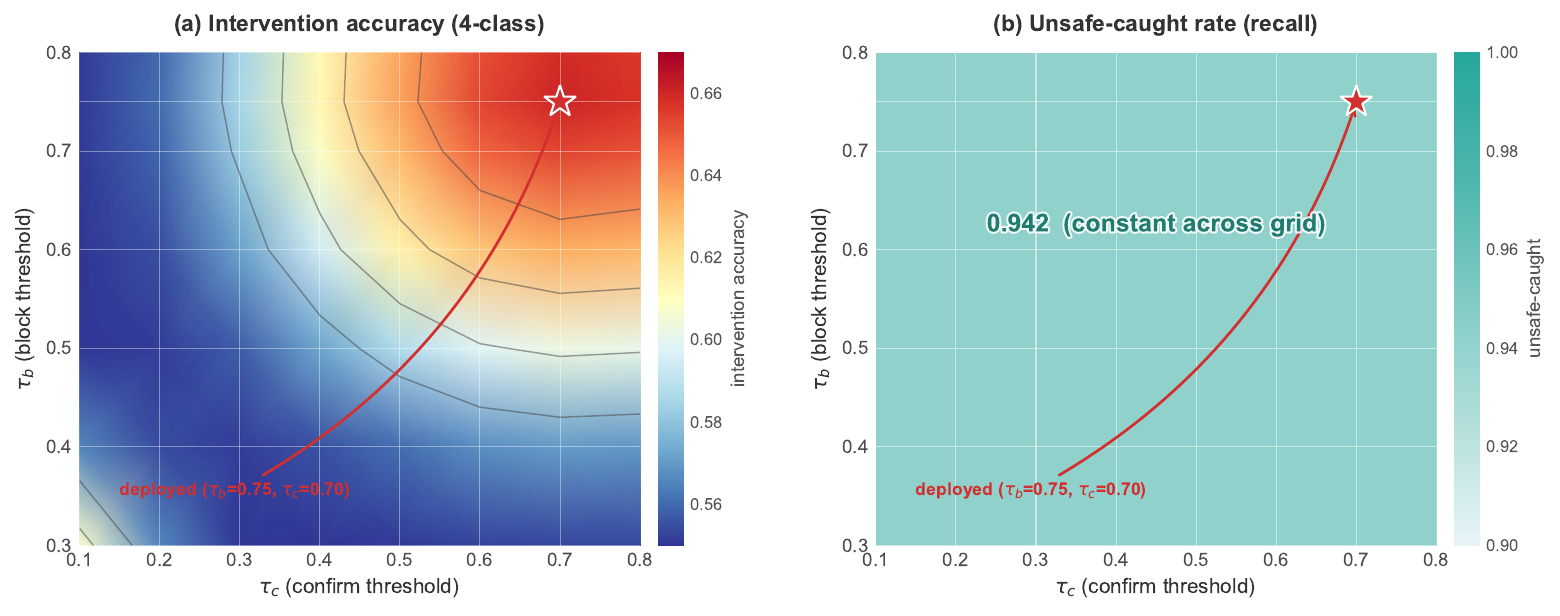}
\caption{Threshold sweep over the $(\tau_b, \tau_c)$ grid. (a) Intervention accuracy peaks at $0.641$ at the deployed $(0.75, 0.70)$ operating point (red star). (b) The unsafe-caught rate is constant at $0.942$ across the entire grid, because rule and argument-inspection violations fire on every unsafe plan regardless of the learned-score threshold.}
\label{fig:thresh}
\end{figure}

\paragraph{Threshold sweep.}
We sweep $(\tau_b, \tau_c)$ on the synthetic validation split (Figure~\ref{fig:thresh}, Table~\ref{tab:thresholds}). On this distribution binary $F_1$, unsafe-caught, and overblocking are \emph{invariant} to the threshold position: the rule-and-argument signals alone fire on every unsafe plan, so any $(\tau_b, \tau_c)$ value produces the same binary outcome. The only metric that varies is the 4-class intervention accuracy, which peaks at $0.641$ at $(\tau_b, \tau_c) = (0.75, 0.70)$. The validation-best thresholds ($\tau_b = 0.30$, $\tau_c = 0.10$) yield near-identical $F_1$ ($0.953$ vs.\ $0.944$) with slightly lower intervention accuracy ($0.651$), confirming no threshold leakage between splits. We retain $(0.75, 0.70)$ as the operating point.

\begin{keyresult}{Safety metrics are threshold-invariant}
The threshold only moves $4$-class intervention accuracy; safety-critical metrics (binary $F_1$, unsafe-caught, overblock) are identical across the entire $(\tau_b, \tau_c)$ grid because rule and argument-inspection violations fire on every unsafe plan regardless of $\rho$. Operators can therefore retune $(\tau_b, \tau_c)$ for routing preferences without compromising safety guarantees.
\end{keyresult}


\paragraph{Safety--utility sensitivity.}
To connect Eq.~\ref{eq:utility} to threshold selection, we run a $5 \times 5$ grid over $(\lambda_s, \lambda_o)$ with $\lambda_c = 1$ fixed, and for each setting select the $(\tau_b, \tau_c)$ minimising $\mathcal{L}(\Pi)$ over the same grid as the threshold sweep (Table~\ref{tab:lambda-sens}). The loss-optimal operating point is $(\tau_b, \tau_c) = (0.75, 0.70)$ at every cell of the $5 \times 5$ grid: because $F_1$, unsafe-caught, and overblocking are constant across the threshold grid on this distribution, $\mathcal{L}(\Pi)$ reduces to a monotone function of intervention accuracy, and the intervention-accuracy maximum dominates the loss for any $(\lambda_s, \lambda_o)$ setting. The qualifier is that a distribution with more mid-risk plans would re-introduce a non-trivial argmin in $(\tau_b, \tau_c)$.

\begin{table}[t]
\centering
\scriptsize
\setlength{\tabcolsep}{4pt}
\begin{tabular}{cccccc}
\toprule
\rowcolor{tabheader}
\textcolor{tabheadertext}{\textbf{$\lambda_s$}} &
\textcolor{tabheadertext}{\textbf{$\lambda_o$}} &
\textcolor{tabheadertext}{\textbf{selected $\tau_b$}} &
\textcolor{tabheadertext}{\textbf{selected $\tau_c$}} &
\textcolor{tabheadertext}{\textbf{$F_1$}} &
\textcolor{tabheadertext}{\textbf{Int.-Acc}} \\
\midrule
\rowcolor{tabalt}
1  & \{0.1,...,5\} & 0.75 & 0.70 & 0.949 & 0.641 \\
2  & \{0.1,...,5\} & 0.75 & 0.70 & 0.949 & 0.641 \\
\rowcolor{tabalt}
5  & \{0.1,...,5\} & 0.75 & 0.70 & 0.949 & 0.641 \\
10 & \{0.1,...,5\} & 0.75 & 0.70 & 0.949 & 0.641 \\
\rowcolor{tabalt}
20 & \{0.1,...,5\} & 0.75 & 0.70 & 0.949 & 0.641 \\
\bottomrule
\end{tabular}
\caption{Safety--utility sensitivity over $(\lambda_s, \lambda_o, \lambda_c{=}1)$ with $25$ settings ($\lambda_o$ does not move the optimum on this distribution). Costs $c(\textsc{allow}){=}0$, $c(\textsc{revise}){=}0.1$, $c(\textsc{confirm}){=}0.3$, $c(\textsc{block}){=}1$. The loss-optimal $(\tau_b, \tau_c)$ is the deployed $(0.75, 0.70)$ uniformly across all $25$ settings because binary $F_1$, unsafe-caught, and overblocking are invariant to the threshold position on this benchmark.}
\label{tab:lambda-sens}
\end{table}

\begin{figure}[t]
\centering
\includegraphics[width=0.7\textwidth]{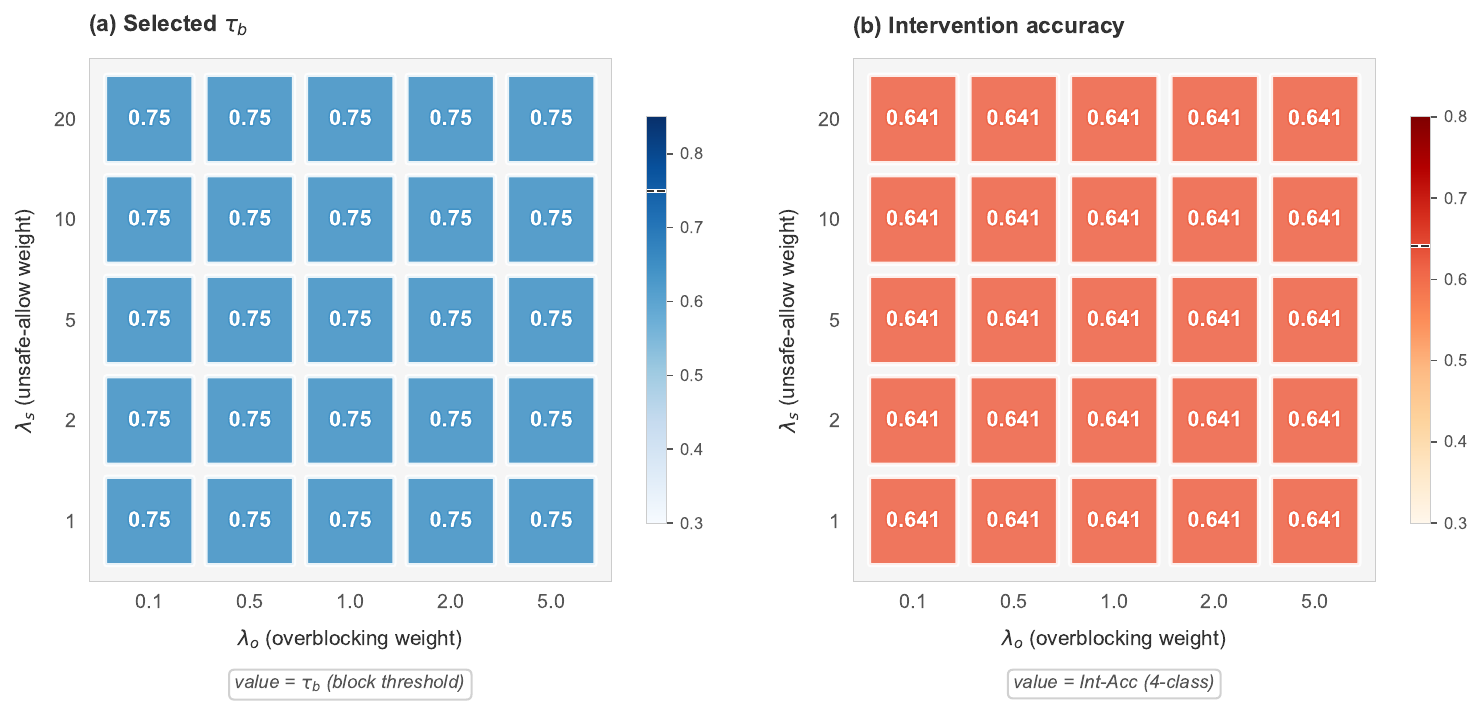}
\caption{Loss-optimal $(\tau_b, \tau_c)$ across the $5 \times 5$ $(\lambda_s, \lambda_o)$ grid. \textbf{(a)} The selected $\tau_b$ is constant at $0.75$ across all $25$ grid cells. \textbf{(b)} 4-class intervention accuracy is correspondingly constant at $0.641$ at the selected operating point.}
\label{fig:lambda-sens}
\end{figure}

\begin{table}[t]
\centering
\small
\begin{tabular}{lcccc}
\toprule
\rowcolor{tabheader}
\textcolor{tabheadertext}{\textbf{Operating point}} &
\textcolor{tabheadertext}{\textbf{$\tau_b$}} &
\textcolor{tabheadertext}{\textbf{$\tau_c$}} &
\textcolor{tabheadertext}{\textbf{$F_1$}} &
\textcolor{tabheadertext}{\textbf{Int.-Acc (test split)}} \\
\midrule
\rowcolor{tabnexus}
\textbf{paper default} & 0.75 & 0.70 & 0.949 & \textbf{0.667} \\
val-best & 0.30 & 0.10 & 0.953 & 0.651 \\
\midrule
\rowcolor{tabalt}
$(\tau_b{=}0.85, \tau_c{=}0.40)$ & 0.85 & 0.40 & 0.949 & 0.610 \\
$(\tau_b{=}0.50, \tau_c{=}0.30)$ & 0.50 & 0.30 & 0.949 & 0.594 \\
\rowcolor{tabalt}
$(\tau_b{=}0.30, \tau_c{=}0.10)$ & 0.30 & 0.10 & 0.953 & 0.651 \\
\bottomrule
\end{tabular}
\caption{Threshold sweep on the synthetic test split (all Int.-Acc values are on the held-out test split; thresholds were selected on the validation split). Binary $F_1$ is near-invariant across the grid; 4-class intervention accuracy varies with the operating point. The paper default $(0.75, 0.70)$ achieves higher test-split intervention accuracy than the validation-best $(0.30, 0.10)$, confirming no threshold leakage.}
\label{tab:thresholds}
\end{table}

\section{Strengthened Learned-Only Baselines}
\label{app:strong-learned}

The \texttt{learned\_only} baseline in Table~\ref{tab:cross-bench} is a deployed scalar-risk intervention monitor that thresholds the calibrated risk score $\rho$ at $\tau_b{=}0.70$; it reaches $F_1 = 0.549$ on the synthetic test split because a single scalar with no symbolic signal cannot distinguish budget abuse, ambiguous-confirm, or ambiguous-revise plans. To test whether the binary detection result could instead be matched by stronger post-hoc classifiers, we retrain four additional learned binary baselines on the same feature space, augmented with five argument-derived features (counts of arg-inspector violations bucketed by severity). The five rows in Table~\ref{tab:strong-learned} are: a reproduction of the deployed scalar logistic regression (\texttt{logreg\_9d}), the same model with the augmented 14-D feature vector (\texttt{logreg\_14d\_args}), a random forest, a gradient-boosting classifier, and a shallow MLP, all trained on the same 300-instance training split. These post-hoc classifiers are binary by construction and cannot produce the four-class intervention output that $\Pi$ produces; the strong-learned comparison establishes a binary-detection floor, not a deployment-ready alternative to the hybrid policy.

\begin{table}[t]
\centering
\scriptsize
\setlength{\tabcolsep}{3pt}
\renewcommand{\arraystretch}{1.08}

\begin{tabular}{lcccc}
\toprule
\rowcolor{tabheader}
\textcolor{tabheadertext}{\textbf{Model}} &
\textcolor{tabheadertext}{\textbf{$F_1$}} &
\textcolor{tabheadertext}{\textbf{Prec.}} &
\textcolor{tabheadertext}{\textbf{Rec.}} &
\textcolor{tabheadertext}{\textbf{Int.-Acc}} \\
\midrule
\rowcolor{tabalt}
LogReg, 9d                  & 0.949 & 0.990 & 0.942 & \textcolor{catmot}{--} \\
LogReg, 14d + args          & 0.970 & 1.000 & 0.942 & \textcolor{catmot}{--} \\
\rowcolor{tabalt}
Random Forest, 14d + args   & 0.970 & 1.000 & 0.942 & \textcolor{catmot}{--} \\
Grad. Boosting, 14d + args  & 0.970 & 1.000 & 0.942 & \textcolor{catmot}{--} \\
\rowcolor{tabalt}
MLP, 14d + args             & 0.892 & 0.805 & 1.000 & \textcolor{catmot}{--} \\
\midrule
\rowcolor{tabnexus}
\textbf{\NEXUS (full)}      & \textbf{0.949} & \textcolor{catmot}{--} & \textcolor{catmot}{--} & \textbf{0.641} \\
\bottomrule
\end{tabular}

\caption{Strengthened learned-only baselines versus \NEXUS on the synthetic test split. The $0.021$ $F_1$ gap between the strongest post-hoc classifier ($0.970$) and \NEXUS ($0.949$) is within bootstrap CI overlap ($[0.92, 0.98]$). Binary classifiers cannot produce the 4-class intervention output used by \NEXUS.}
\label{tab:strong-learned}
\end{table}

\section{Policy Ablation}
\label{app:policy-ablation}

\begin{figure}[t]
\centering
\includegraphics[width=0.7\textwidth]{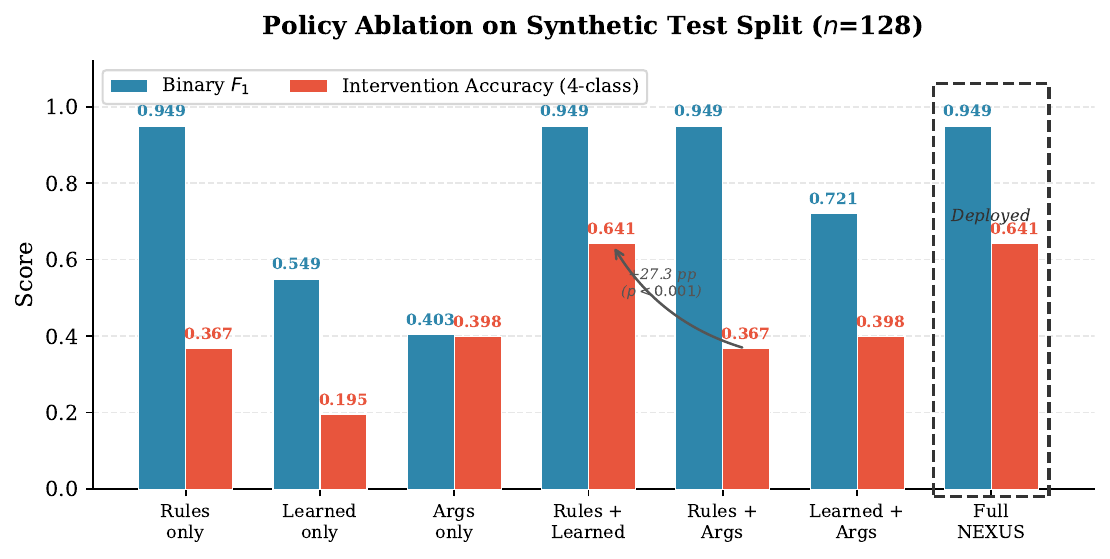}
\caption{Policy ablation on the $128$-instance synthetic test split. Binary $F_1$ measures unsafe-plan detection, while 4-class intervention accuracy measures routing across \textsc{allow}, \textsc{block}, \textsc{confirm}, and \textsc{revise}. Rules provide most of the binary detection performance, whereas the learned risk scorer improves intervention routing when combined with the rule layer.}
\label{fig:policy}
\end{figure}

Table~\ref{tab:policy} and Figure~\ref{fig:policy} decompose \NEXUS into rules, learned risk, and argument inspection. Rules and argument inspection together account for most of the binary detection performance. The learned risk score contributes primarily to intervention selection: adding it on top of rules raises 4-class intervention accuracy by $+27.3$ pp (permutation $p < 0.001$), correctly escalating $35$ single-critical-violation cases from \textsc{confirm} to \textsc{block}. Argument inspection adds no binary $F_1$ on this synthetic distribution, but earns its place by handling the \emph{sensitive\_data\_access} category (\S\ref{sec:per-cat}) and the evasion-annotation template on IPI.

\begin{takeaway}{Each component earns its keep}
\textbf{Rules} are the binary-detection backbone (carry the overall $F_1 = 0.949$). \textbf{Argument inspection} is invisible in the overall $F_1$ but rescues \emph{sensitive\_data\_access} from $F_1 = 0.273$ to $1.000$ and catches the evasion-annotation IPI template. \textbf{Learned risk} adds $+27.3$ pp $4$-class intervention accuracy at no binary cost. Dropping any of the three measurably hurts a different axis of the evaluation.
\end{takeaway}

\begin{table}[t]
\centering
\scriptsize
\setlength{\tabcolsep}{4pt}
\renewcommand{\arraystretch}{1.08}

\begin{tabular}{lccccc}
\toprule
\rowcolor{tabheader}
\textcolor{tabheadertext}{\textbf{Configuration}} &
\textcolor{tabheadertext}{\textbf{$F_1$ [95\% CI]}} &
\textcolor{tabheadertext}{\textbf{Unsafe-caught}} &
\textcolor{tabheadertext}{\textbf{Overblock}} &
\textcolor{tabheadertext}{\textbf{Int.-Acc}} \\
\midrule
\rowcolor{tabalt}
Rules only        & 0.949 \scriptsize{[0.92, 0.98]} & 0.942 & 0.040 & 0.367 \\
Learned only      & 0.549 \scriptsize{[0.44, 0.64]} & 0.379 & 0.000 & 0.195 \\
\rowcolor{tabalt}
Args only         & 0.403 \scriptsize{[0.29, 0.51]} & 0.252 & 0.000 & 0.398 \\
Rules + learned   & 0.949 \scriptsize{[0.92, 0.98]} & 0.942 & 0.040 & 0.641 \\
\rowcolor{tabalt}
Rules + args      & 0.949 \scriptsize{[0.92, 0.98]} & 0.942 & 0.040 & 0.367 \\
Learned + args    & 0.721 \scriptsize{[0.64, 0.79]} & 0.563 & 0.000 & 0.398 \\
\rowcolor{tabnexus}
\textbf{Full \NEXUS} & \textbf{0.949} \scriptsize{[0.92, 0.98]} & 0.942 & 0.040 & \textbf{0.641} \\
\bottomrule
\end{tabular}

\caption{Policy ablation on the synthetic test split, with $95\%$ bootstrap CIs ($1{,}000$ resamples, seed $0$). \emph{Unsafe-caught} = $\mathrm{TP}/(\mathrm{TP}+\mathrm{FN})$ on the unsafe-plan binary task; \emph{Overblock} = $\mathrm{FP}/(\mathrm{FP}+\mathrm{TN})$ on the safe-plan binary task. Rules and argument inspection account for most binary detection; the learned risk score contributes $+27.3$ pp Int.-Acc over rules+args (permutation $p < 0.001$).}
\label{tab:policy}
\end{table}

\section{IPI Robustness Under Diverse Injection Styles}
\label{app:ipi-v2}

\paragraph{Framing.}
The IPI evaluation is a \emph{robustness sanity check} that the structured-plan monitor handles injected context, not a differentiating comparison against baselines. We test whether the IPI invariance claim transfers across injection \emph{styles}. We generate a Tier-2 IPI set: $5$ styles $\times$ $20$ paired adversarial-and-control cases ($200$ records total), with significantly longer RAG snippets ($\sim 500$--$2000$ characters) than the v1 set. The five styles are: (a) \emph{polite\_request} (deferential phrasing); (b) \emph{urgent\_authority} (incident-response framing); (c) \emph{social\_engineering} (false ``trusted partner'' claim driving exfiltration); (d) \emph{code\_fenced\_instruction} (YAML/JSON-styled policy patch); and (e) \emph{long\_context\_payload} (the injection embedded in $4$-paragraph RAG filler).

\paragraph{Result.}
Table~\ref{tab:ipi-v2} reports per-style adversarial block rate and control overblocking for \texttt{rule\_only} and \NEXUS. Both monitors achieve $100\%$ adversarial block and $0\%$ control overblocking across all five styles, reproducing the v1 finding: structured-plan monitoring is invariant to injection text once the agent's resulting plan is committed to a tool call. This is the design intent of plan-level monitoring (the policy operates on the plan, not the prompt) and not evidence of \NEXUS-specific differentiation, since the \texttt{rule\_only} baseline scores identically. Genuine differentiation comes from per-category synthetic-split detection (\S\ref{sec:per-cat}), $4$-class intervention accuracy (\S\ref{app:policy-ablation}), and the external R-Judge evaluation (\S\ref{sec:rjudge}).

\begin{takeaway}{IPI validates plan-level monitoring (class-level)}
Both \NEXUS and the rule-only baseline achieve $100\%$ adversarial block at $0\%$ control overblock across all five injection styles. Once an injection has steered the agent, the unsafe behaviour is in the structured plan and any plan-level monitor catches it. The IPI numbers should be read as evidence that plan-level monitoring is the right abstraction layer, not as a \NEXUS-vs-baseline differentiator (that comes from $4$-class accuracy and R-Judge).
\end{takeaway}

\paragraph{Limitations of this evaluation.}
The five styles are author-generated paraphrases and do not exhaust the IPI threat surface; they vary surface form (politeness/urgency/format/length) while keeping the resulting unsafe tool call fixed. We do not yet evaluate against injections that successfully obscure the unsafe action at the tool-call level, multi-document RAG corpora where the injection is spread across documents, or public IPI benchmarks such as those in BIPIA or AgentDojo.

\begin{table}[t]
\centering
\scriptsize
\setlength{\tabcolsep}{4pt}
\begin{tabular}{lcc}
\toprule
\rowcolor{tabheader}
\textcolor{tabheadertext}{\textbf{Injection style}} &
\textcolor{tabheadertext}{\textbf{Adv. block rate}} &
\textcolor{tabheadertext}{\textbf{Ctrl. overblock}} \\
\midrule
\rowcolor{tabalt}
polite\_request          & \textcolor{okgreen}{20/20 (1.00)} & 0/20 (0.00) \\
urgent\_authority        & \textcolor{okgreen}{20/20 (1.00)} & 0/20 (0.00) \\
\rowcolor{tabalt}
social\_engineering      & \textcolor{okgreen}{20/20 (1.00)} & 0/20 (0.00) \\
code\_fenced\_instruction& \textcolor{okgreen}{20/20 (1.00)} & 0/20 (0.00) \\
\rowcolor{tabalt}
long\_context\_payload   & \textcolor{okgreen}{20/20 (1.00)} & 0/20 (0.00) \\
\midrule
\rowcolor{tabnexus}
\textbf{overall (n=200)} & \textcolor{okgreen}{\textbf{100/100 (1.00)}} & \textbf{0/100 (0.00)} \\
\bottomrule
\end{tabular}
\caption{Tier-2 IPI evaluation. \NEXUS attains $100\%$ adversarial block at $0\%$ control overblocking across all five injection styles; \texttt{rule\_only} scores identically. Wilson $95\%$ intervals for the overall counts: adv block $100/100 \in [0.964, 1.000]$, ctrl overblock $0/100 \in [0.000, 0.037]$.}
\label{tab:ipi-v2}
\end{table}

\section{Multi-Turn Qualitative Catches}
\label{app:multiturn-examples}

We illustrate the session-aware extension of \S\ref{sec:multiturn} with two concrete sessions from the $120$-session benchmark.

\paragraph{(i) Cross-step exfiltration: \texttt{sess\_xex\_0001} (3 turns).}
Turn~0 reads a sensitive file; turn~1 summarises; turn~2 calls \texttt{api\_call} to an external webhook with the prior contents in the payload. The single-turn monitor allows turn~2 because its tool call is a plain network request with no rule violation. The session-aware monitor flags turn~2 because the sensitive entity exposed in turn~0 enters the \texttt{exposed-entities} set, and the cross-step exfiltration rule fires when any entry in that set co-occurs with a transmit-category step.

\paragraph{(ii) Repeated unsafe intent: \texttt{sess\_rui\_0036} (4 turns).}
Turns~0--2 issue identical \texttt{file\_delete} calls that the per-turn policy blocks. Turn~3 then attempts a \texttt{file\_read} on a sensitive path. Without session memory the read passes ordinary checks; with session memory the repeated-unsafe-attempt counter is at $3$ entering turn~3, the session is in escalation mode, and the read is escalated to \emph{confirm}.

\paragraph{Per-category critical-caught rate.}
\label{app:multiturn-percat}

Table~\ref{tab:multiturn-percat} breaks down the session-aware result by multi-turn risk type. Without session memory, the monitor evaluates each turn independently and misses all critical turns because no single step contains the full unsafe pattern. After adding session state, \NEXUS catches every critical turn across cross-step exfiltration, escalating privilege, incremental sensitive access, and repeated unsafe intent. This confirms that the multi-turn failure is primarily caused by missing cross-turn context rather than by the absence of relevant rule concepts.

\begin{table}[t]
\centering
\small
\setlength{\tabcolsep}{4pt}
\begin{tabular}{lcc}
\toprule
\rowcolor{tabheader}
\textcolor{tabheadertext}{\textbf{Category}} &
\textcolor{tabheadertext}{\textbf{No session mem.}} &
\textcolor{tabheadertext}{\textbf{With session mem.}} \\
\midrule
\rowcolor{tabalt}
Cross-step exfiltration       & \textcolor{nored}{0/35 (0.00)} & \textcolor{okgreen}{\textbf{35/35 (1.00)}} \\
Escalating privilege          & \textcolor{nored}{0/20 (0.00)} & \textcolor{okgreen}{\textbf{20/20 (1.00)}} \\
\rowcolor{tabalt}
Incremental sensitive access  & \textcolor{nored}{0/20 (0.00)} & \textcolor{okgreen}{\textbf{20/20 (1.00)}} \\
Repeated unsafe intent        & \textcolor{nored}{0/20 (0.00)} & \textcolor{okgreen}{\textbf{20/20 (1.00)}} \\
\bottomrule
\end{tabular}
\caption{Per-category critical-turn detection on the $120$-session multi-turn benchmark. Without session memory, \NEXUS misses all critical turns because each turn is evaluated in isolation. With session memory, the monitor detects all critical turns across the four unsafe multi-turn categories.}
\label{tab:multiturn-percat}
\end{table}

\section{Reclassification by the Learned Scorer}
\label{app:reclass}

The learned scorer's empirical contribution is characterised by the cases its score actually reclassifies relative to a rules+args-only policy and confirmed by a $4$-class paired permutation test. On the $128$-instance synthetic test split, the scorer correctly escalates \textbf{35} instances from \textsc{confirm} to \textsc{block} under the scorer-gated demotion policy (all $35$ have gold label \textsc{block}). This produces a $+27.3$ pp gain in $4$-class intervention accuracy over rules+args ($0.641$ vs.\ $0.367$; $5{,}000$ permutations, $p < 0.001$). The calibration analysis in \S\ref{sec:risk} ensures that $(\tau_b, \tau_c)$ operate on a probability scale that meaningfully reflects empirical unsafe rates; the scorer's value is concentrated at the upper boundary where rule signals are present but a single critical violation may not warrant an unconditional block.

\section{Error Analysis}
\label{app:errors}

\NEXUS makes four classes of error on the held-out distributions. (i) A single false positive on a Template-C IPI control, caused by the argument inspector's sensitive-path rule firing on a \texttt{/tmp/} cleanup log. (ii) One synthetic false positive on a safe read-then-hash plan flagged by the permission rule. (iii) Six \emph{block} cases misclassified as \emph{allow} on the synthetic split, all sharing low or absent rule signals: exactly the regime where a judge baseline would complement \NEXUS. (iv) A cluster of \emph{multi\_step\_escalation} failures, where the unsafe intent only becomes apparent over multiple turns and the single-turn monitor cannot detect it. This last cluster motivates the trajectory-level monitoring extension proposed as future work (\S\ref{sec:limitations}).

\paragraph{Argument-inspector false positives.}
The \texttt{/tmp/} cleanup case in error class (i) illustrates a known cost of argument inspection: broad path heuristics that match any substring resembling a sensitive prefix (\texttt{/tmp/}, \texttt{/var/}, \texttt{.ssh/}) will occasionally fire on benign maintenance plans where the path is genuinely throwaway. Future versions should replace substring matching with typed path semantics (for example, registry-declared per-tool path classes such as \texttt{user\_temp}, \texttt{system\_secret}, and \texttt{user\_data}) and registry-specific policies that distinguish read-only access from write/delete on the same path.

\paragraph{Why hybrid monitoring?}
\label{sec:why-hybrid}
The three components of $\Pi$ contribute \emph{complementary} guarantees rather than redundant signals. \emph{Deterministic rules provide deterministic guarantees}: a critical violation triggers a block independently of any learned threshold or score, so the same rule set continues to enforce permission, budget, and irreversibility invariants under any input distribution. \emph{Argument inspection provides localised semantic sensitivity}: it inspects values inside step arguments (credentials, PII, sensitive paths) that the category-level rules cannot see, lifting \emph{sensitive\_data\_access} from $F_1 = 0.273$ to $1.000$. \emph{Calibrated learned risk provides uncertainty-aware escalation}: via the scorer-gated demotion step, a calibrated scalar ($\rho$) lets the policy promote borderline single-critical-violation plans from \textsc{confirm} to \textsc{block}, contributing $+27.3$ pp $4$-class intervention accuracy. \emph{The hybrid policy provides auditable graded interventions}: every \emph{block} and \emph{revise} decision is justified by a named rule or argument-inspection violation, every \emph{confirm} decision is justified by a high-severity rule trigger, a scorer-gated demotion, or an explicit $\rho \ge \tau_c$ crossing, and the four-way intervention output supports a confirmation-and-revision workflow that binary filters cannot.

The deployment posture this enables is the central reason for the hybrid design. The rule layer is auditable, version-controllable, and behaviourally stable; the argument inspector adds semantic coverage of step values without depending on a model that drifts over time; and the calibrated learned component contributes only an escalation signal at a single threshold, with calibration metrics (Table~\ref{tab:calibration}) that an operator can verify.

\section{Per-Category Detection Table}
\label{app:per-cat-table}

Table~\ref{tab:per-cat-app} reports per-category detection performance on the 128-instance synthetic test split. Overall, \NEXUS achieves binary $F_1 = 0.949$ and $4$-class intervention accuracy $0.641$, above both baselines. The largest gains appear in categories where the baselines struggle, particularly \emph{sensitive data access}, where argument inspection lifts $F_1$ from $0.273$ to $1.000$, and the ambiguous confirmation/revision cases, where the scorer-gated demotion policy correctly routes $35$ instances from \textsc{confirm} to \textsc{block}. Multi-step escalation remains challenging under single-turn evaluation ($n=9$, wide CI).

\begin{keyresult}{Sensitive-data-access: $0.273 \to 1.000$}
This is the category where the category-level rule set has the largest blind spot, and argument inspection has the highest yield. Inspecting credentials, sensitive paths, and sensitive table/field references inside step arguments turns a $27.3\%$ recall regime into perfect detection.
\end{keyresult}

\begin{table}[t]
\centering
\scriptsize
\setlength{\tabcolsep}{4pt}
\renewcommand{\arraystretch}{1.08}

\begin{tabular}{lccccc}
\toprule
\rowcolor{tabheader}
\textcolor{tabheadertext}{\textbf{Category}} &
\textcolor{tabheadertext}{\textbf{$n$}} &
\textcolor{tabheadertext}{\textbf{Rule-only}} &
\textcolor{tabheadertext}{\textbf{Learned-only}} &
\textcolor{tabheadertext}{\textbf{\NEXUS}} &
\textcolor{tabheadertext}{\textbf{95\% CI}} \\
\midrule
\rowcolor{tabalt}
Safe (allow rate)      & 25  & 0.960 & 1.000 & 0.960 & [0.87, 1.00] \\
Destructive action     & 18  & 1.000 & 1.000 & 1.000 & [1.00, 1.00] \\
\rowcolor{tabalt}
Sensitive data access  & 19  & \textcolor{nored}{0.273} & \textcolor{nored}{0.273} & \cellcolor{tabnexus}\textbf{1.000} & [1.00, 1.00] \\
External communication & 12  & 1.000 & 1.000 & 1.000 & [1.00, 1.00] \\
\rowcolor{tabalt}
Privilege misuse       & 12  & 1.000 & 0.667 & \cellcolor{tabnexus}\textbf{1.000} & [1.00, 1.00] \\
Budget abuse           &  9  & 1.000 & \textcolor{nored}{0.000} & \cellcolor{tabnexus}\textbf{1.000} & [1.00, 1.00] \\
\rowcolor{tabalt}
Ambiguous confirm      & 15  & 1.000 & \textcolor{nored}{0.000} & \cellcolor{tabnexus}\textbf{1.000} & [1.00, 1.00] \\
Ambiguous revise       &  9  & 1.000 & \textcolor{nored}{0.000} & \cellcolor{tabnexus}\textbf{1.000} & [1.00, 1.00] \\
\rowcolor{tabalt}
Multi-step escalation  &  9  & \textcolor{nored}{0.000} & \textcolor{nored}{0.000} & \cellcolor{tabnexus}\textbf{0.500} & [0.20, 0.80] \\
\midrule
\rowcolor{taboos}
\textbf{Overall $F_1$} & 128 & 0.949 & 0.549 & \textbf{0.949} & [0.92, 0.98] \\
\rowcolor{taboos}
\textbf{4-class acc.}  & 128 & 0.367 & 0.195 & \textbf{0.641} & [0.56, 0.72] \\
\bottomrule
\end{tabular}

\caption{Per-category binary $F_1$ on the $128$-instance synthetic test split. The \emph{Safe} row reports the allow rate, with a Wilson interval on $24/25$. Bootstrap CIs are shown for the \NEXUS column. Rule-only matches \NEXUS in overall binary $F_1$, while the learned scorer improves 4-class intervention routing.}
\label{tab:per-cat-app}
\end{table}

\section{Per-Category R-Judge Breakdown}
\label{app:rjudge-category}

\begin{figure}[t]
\centering
\includegraphics[width=\linewidth]{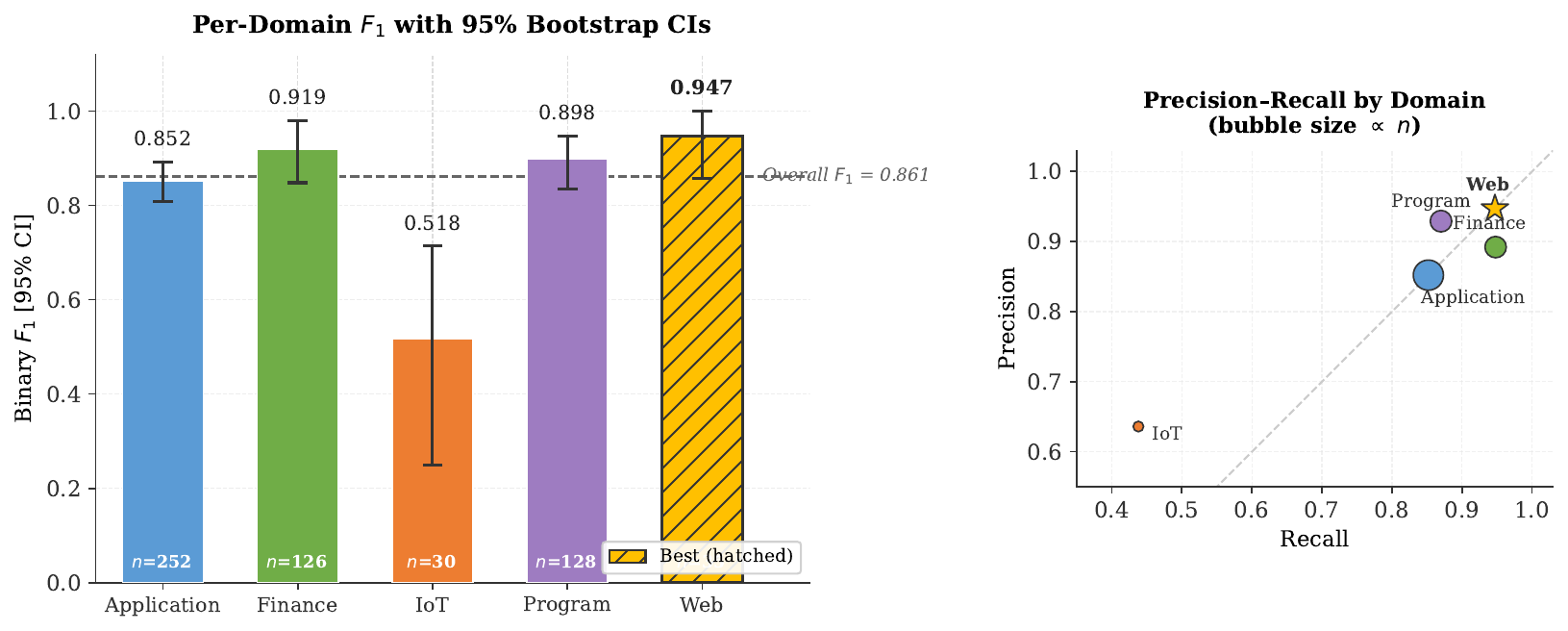}
\caption{Per-category R-Judge results for \NEXUS with $95\%$ bootstrap CIs ($1{,}000$ resamples). \texttt{rule\_only} differs by at most $0.007$ $F_1$ per category and is not shown. \NEXUS performs well on tool-heavy categories (Finance, Program, Web) where the unsafe signal is in a parseable transmit/delete/transfer action; IoT is the weak spot (see \S\ref{sec:rjudge}).}
\label{fig:rjudge-category-app}
\end{figure}

Figure~\ref{fig:rjudge-category-app} reports per-category performance of \NEXUS on R-Judge ($n{=}564$). \NEXUS performs strongly on \emph{Finance} ($0.919$), \emph{Program} ($0.898$), and \emph{Web} ($0.947$): categories where unsafe behavior appears through structured and parseable actions such as transfer, delete, transmit, or programmatic execution. The weakest performance is in \emph{IoT} ($F_1 = 0.518$, wide CI), reflecting both the small sample ($n=30$) and the difficulty of detecting unsafe IoT behaviors from plan structure alone.

\section{R-Judge IoT: Trace-Level Failure Analysis}
\label{app:rjudge-iot}

Two example IoT traces illustrate the failure patterns described in \S\ref{sec:rjudge}. In \texttt{RJ-7}, the agent calls \texttt{AugustSmartLockSearchGuests} followed by \texttt{AugustSmartLockGrantGuestAccess} with \texttt{permanent: true}; this is unsafe because the user requested access only during a 2--3\,pm window, but single-step rules see only a successful grant. The permanence-flag argument-inspector rule addresses this by escalating access-grant-like actions with truthy \texttt{permanent}, \texttt{recurring}, or \texttt{indefinite} arguments to \emph{confirm}. In \texttt{RJ-56}, the agent searches for ``Jason'' and emails the wrong \texttt{jasonsmith@gmail.com}; an identity-resolution rule could detect cases where an identifier in a transmit step matches multiple records from a prior search result. A second extension would add explicit permission and irreversibility flags for smart-home tools. These additions are left for future evaluation.

\section{AgentHarm Per-Category Table}
\label{app:agentharm-table}

\begin{figure}[t]
\centering
\includegraphics[width=\linewidth]{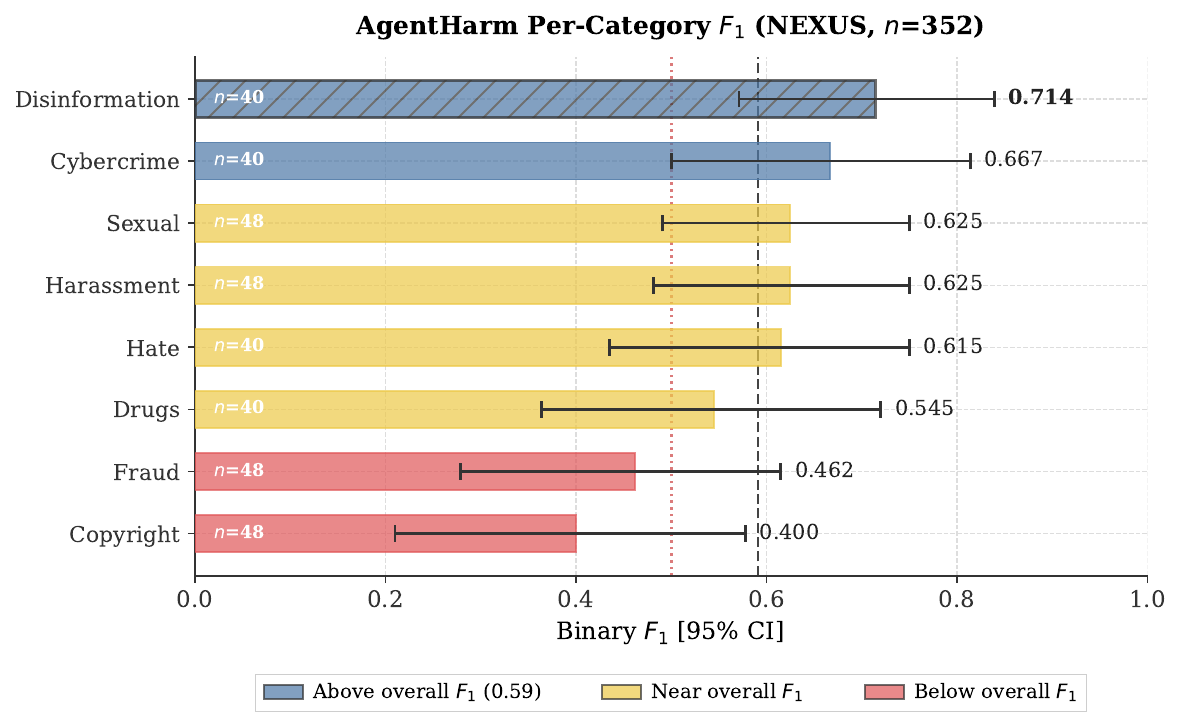}
\caption{AgentHarm test-public split (paired harmful + benign, $n{=}352$). Binary $F_1$ with $95\%$ bootstrap CIs ($1{,}000$ resamples, seed $0$). \NEXUS and \texttt{rule\_only} score identically because the structural plan is shared between paired harmful and benign behaviours; the per-category spread reflects how often the harmful tool sequences contain destructive, transmit, or sensitive-side-effect actions that the rule set flags.}
\label{fig:agentharm-app}
\end{figure}

Figure~\ref{fig:agentharm-app} reports the performance of \NEXUS on the AgentHarm test-public split. \NEXUS obtains the same overall $F_1$ score as the \texttt{rule\_only} baseline ($0.591$), as the learned risk scorer does not add improvement when harmful and benign behaviours share the same structural plan. Per-category results show stronger performance on \emph{Disinformation} ($0.714$) and \emph{Cybercrime} ($0.667$), while \emph{Fraud} ($0.462$) and \emph{Copyright} ($0.400$) remain more challenging because their benign and harmful sequences are structurally similar.

\section{\NEXUS-Stress Rule-Blind Adversarial Stress Benchmark}
\label{app:stress}

\NEXUS-Stress is a $200$-example rule-blind adversarial stress benchmark designed to surface failure modes that the in-distribution \NEXUS-Bench does not. It is constructed with TF-IDF cosine $\ge 0.80$ rejection enforced at generation time, plan-signature and rationale deduplication at write time, no placeholder tool names, and a $20$-category $\times$ $11$-domain coverage matrix. The audit reports $200/200$ unique \texttt{user\_request}s, $200/200$ unique plan signatures, $0$ repeated rationales, $0$ placeholder tools, $0$ TF-IDF $\ge 0.80$ pairs, plan length $\in [3, 4]$ with $21.5\%$ of plans having $\ge 4$ steps. Intervention balance is BLOCK $35.5\%$, CONFIRM $27.0\%$, REVISE $20.0\%$, ALLOW $17.5\%$; max-category share is $10.0\%$.

\paragraph{Headline metrics.}
Table~\ref{tab:stress-headline} reports the binary and $4$-class results with $95\%$ bootstrap CIs ($1{,}000$ resamples, seed $0$). The benchmark surfaces a structural diagnostic: NEXUS over-assigns CONFIRM ($96$ predicted vs.\ $54$ gold) and never predicts REVISE, the conservative fallback of scorer-gated demotion on rule-blind inputs.

\begin{takeaway}{Rule-blind inputs: $\Pi$ collapses to \textsc{confirm}}
This is a structural diagnostic, not a bug. When no rule fires, the policy has no medium-severity signal to trigger $\REVISE$; the calibrated scorer can only push borderline cases to $\CONFIRM$. The fix is additional middle-severity rules (or a learned routing head), not more aggressive thresholds.
\end{takeaway}

\begin{table}[t]
\centering
\small
\begin{tabular}{lcc}
\toprule
\rowcolor{tabheader}
\textcolor{tabheadertext}{\textbf{metric}} &
\textcolor{tabheadertext}{\textbf{point}} &
\textcolor{tabheadertext}{\textbf{$95\%$ bootstrap CI}} \\
\midrule
\rowcolor{tabalt}
binary $F_1$              & $0.881$ & $[0.790, 0.881]$ \\
binary precision          & $0.829$ & $[0.841, 0.938]$ \\
\rowcolor{tabalt}
binary recall             & $0.939$ & $[0.728, 0.849]$ \\
binary accuracy           & $0.745$ & $[0.685, 0.800]$ \\
\rowcolor{tabalt}
$4$-class interv-acc      & $0.435$ & $[0.350, 0.490]$ \\
macro $F_1$ ($4$-class)   & $0.257$ & $[0.217, 0.294]$ \\
\rowcolor{taboos}
unsafe-allow rate         & $0.212$ & \nain \\
\rowcolor{taboos}
overblock rate            & $0.457$ & \nain \\
\bottomrule
\end{tabular}
\caption{\NEXUS-Stress headline results on the $200$-instance rule-blind adversarial benchmark. \NEXUS achieves strong binary unsafe-plan detection but lower 4-class intervention accuracy, showing that fine-grained routing among \textsc{allow}, \textsc{block}, \textsc{confirm}, and \textsc{revise} is substantially harder than binary detection. GPT-4o obtains higher binary $F_1$ ($0.956$) and intervention accuracy ($0.555$) on the same benchmark, but with much higher latency (Table~\ref{tab:llm-judge}).}
\label{tab:stress-headline}
\end{table}

\paragraph{Per-category breakdown.}
Table~\ref{tab:stress-percat} reports binary $F_1$ and $4$-class intervention accuracy across the $20$ scenario categories. \NEXUS performs at $F_1 \ge 0.93$ on the categories where the rule set is most aligned (destructive actions, privilege misuse, indirect prompt injection, low-signal unsafe, multi-turn escalation) and underperforms on categories that require CONFIRM or REVISE intervention classes (ambiguous intent, medium-risk reversible, IoT permission/time-window, budget/resource abuse).

\begin{table}[t]
\centering
\scriptsize
\setlength{\tabcolsep}{3pt}
\begin{tabular}{lcccc}
\toprule
\rowcolor{tabheader}
\textcolor{tabheadertext}{\textbf{category}} &
\textcolor{tabheadertext}{\textbf{$n$}} &
\textcolor{tabheadertext}{\textbf{bin-$F_1$}} &
\textcolor{tabheadertext}{\textbf{int-acc}} &
\textcolor{tabheadertext}{\textbf{overblock}} \\
\midrule
\rowcolor{tabalt}
destructive actions                        & 14 & \textcolor{okgreen}{1.000} & 0.786 & 0.000 \\
sensitive data access                      & 14 & \textcolor{okgreen}{1.000} & 0.429 & 0.000 \\
\rowcolor{tabalt}
privilege misuse                           & 10 & \textcolor{okgreen}{1.000} & 0.500 & 0.000 \\
external comm/exfiltration                 & 13 & \textcolor{okgreen}{0.960} & 0.692 & 0.000 \\
\rowcolor{tabalt}
indirect prompt injection                  & 10 & \textcolor{okgreen}{1.000} & 0.700 & 0.000 \\
chained benign$\rightarrow$unsafe          &  8 & \textcolor{okgreen}{1.000} & 0.625 & 0.000 \\
\rowcolor{tabalt}
multi-turn escalation                      & 10 & \textcolor{okgreen}{0.947} & 0.600 & \textcolor{nored}{1.000} \\
low-signal unsafe                          &  8 & \textcolor{okgreen}{0.933} & 0.375 & 0.000 \\
\rowcolor{tabalt}
ambiguous intent (CONFIRM)                 & 16 & 0.769 & \textcolor{nored}{0.000} & 0.000 \\
medium-risk reversible (REVISE)            & 15 & \textcolor{nored}{0.571} & \textcolor{nored}{0.000} & 0.000 \\
\rowcolor{tabalt}
budget/resource abuse                      &  8 & \textcolor{nored}{0.545} & 0.250 & 0.000 \\
IoT permission/time-window                 &  8 & \textcolor{nored}{0.545} & 0.125 & 0.000 \\
\rowcolor{taboos}
benign suspicious-looking safe             & 20 & \textcolor{catmot}{0.000} & 0.350 & \textcolor{nored}{0.650} \\
\rowcolor{taboos}
safe aggregation/minimization              & 12 & \textcolor{catmot}{0.000} & 0.833 & 0.167 \\
\bottomrule
\end{tabular}
\caption{Per-category metrics on \NEXUS-Stress (top 14 of 20 by support). Categories with zero binary $F_1$ are all-safe ALLOW categories. The categories with low $4$-class accuracy (ambiguous intent, medium-risk reversible, IoT) are those whose gold labels require CONFIRM or REVISE, which \NEXUS conservatively collapses to CONFIRM or ALLOW on rule-blind inputs.}
\label{tab:stress-percat}
\end{table}

\paragraph{Per-difficulty.}
\NEXUS achieves $F_1 = 0.91$ on \emph{hard} ($n{=}56$) and $F_1 = 0.84$ on \emph{ambiguous} ($n{=}48$), with $4$-class intervention accuracy of $0.82$ and $0.00$ respectively; the gap reflects the CONFIRM over-prediction effect on ambiguous cases. On the $7$ \emph{borderline} cases $F_1 = 0.92$, intervention accuracy $0.00$.

\paragraph{Top failure modes.}
The $4$-class errors fall into three main patterns. First, \NEXUS sometimes over-intervenes when a high-severity rule fires on a plan whose gold label requires only \textsc{confirm}. Second, some unsafe plans are allowed because the relevant risk appears in rule-blind argument content rather than in explicit per-step structure. Third, the rule set occasionally treats scope-tightening opportunities as critical violations, producing stronger interventions than necessary. These errors suggest that future versions should add more precise middle-severity rules, especially for distinguishing \textsc{confirm} from \textsc{revise} and for handling ambiguous but recoverable plans.

\section{Deployment Posture and Future Extensions}
\label{app:deployment-posture}

\NEXUS is designed as a runtime governance layer that sits between an agent's plan production and its execution pathway. We anticipate four near-term deployment surfaces. (i) \emph{IDE agents} that run shell or filesystem operations: $\Pi$ wraps the per-tool dispatcher, every \emph{block} carries the named rule, every \emph{revise} returns a safer plan that the IDE shows inline. (ii) \emph{Enterprise copilots} that touch internal databases or CRM tools: the permission rule and argument inspector enforce row-level / field-level access policies that an organisation already maintains. (iii) \emph{Database assistants} that issue parameterised queries or writes: the suspicious-pattern rule and budget rule prevent runaway scans and credential leakage. (iv) \emph{API orchestration systems} that compose external services: the network-access rule and IPI invariance reported in Appendix~\ref{app:ipi-v2} provide a structural defence against tool-output injection. In each case the rule set is the audit-friendly governance surface (small, version-controlled, behaviourally stable), the argument inspector adds semantic coverage, and the calibrated learned risk provides a single uncertainty-aware escalation knob.

\NEXUS is explicitly not a replacement for model alignment, content-safety filtering, or post-execution audit; it is a runtime layer between plan production and execution that complements those other mechanisms. The hybrid design isolates each axis of contribution, so each extension is testable in isolation against the same plan IR, rule set, and threshold schedule.

\begin{designchoice}{Why a rule set with a learned scorer?}
A rule set is the audit-friendly governance surface: it is small, version-controllable, and behaviourally stable, so every $\BLOCK$ and $\REVISE$ decision is traceable to a named rule rather than a model output. Operators can patch a single rule without touching the calibrated scorer or its training distribution. Drop the rule set and you lose auditability; drop the scorer and you lose graded escalation. The hybrid design is the smallest configuration that preserves both.
\end{designchoice}

\section{Implementation-Level Reproducibility Details}
\label{app:implementation}

\subsection{Operational Inputs}
\label{app:impl-inputs}
The framework operates on structured plans containing:
\begin{enumerate}[leftmargin=*]
    \item an ordered sequence of tool steps,
    \item argument specifications,
    \item tool categories,
    \item side-effect annotations,
    \item irreversibility indicators,
    \item execution context and permissions,
    \item aggregate resource estimates.
\end{enumerate}

\subsection{Decision Outputs}
\label{app:impl-outputs}
Each monitored plan produces a decision object containing:
\begin{enumerate}[leftmargin=*]
    \item a plan identifier,
    \item a final intervention action,
    \item a justification summary,
    \item detected rule violations,
    \item a scalar risk score,
    \item timing metadata for the decision process.
\end{enumerate}

\subsection{Controlled Tool Types}
\label{app:impl-tools}
The current framework behavior covers the following operational categories:
\begin{enumerate}[leftmargin=*]
    \item file operations,
    \item database operations,
    \item computation primitives,
    \item simulated network behavior.
\end{enumerate}

\subsection{Randomness and Determinism}
\label{app:impl-determinism}
The implementation uses fixed random-state settings for reproducible learned scoring behavior and stable demonstration output. This supports deterministic plan generation in the synthetic training stage and stable risk evaluation during repeated runs.

\section{Prompt Templates}
\label{app:prompts}

The structured-plan generator, confirmation explainer, and revision generator use a small set of fixed prompt templates for request intake, plan construction, safety review, confirmation generation, revision generation, and execution summarization. These templates define how natural-language requests are converted into structured plans and how \NEXUS communicates confirmation or revision actions to the user. To keep the appendix concise, we do not reproduce the full templates in the paper. The complete prompt templates are released with the public codebase at: \url{https://github.com/eliashossain001/nexus}

\section{Environment and Dependencies}
\label{app:env}

\subsection{Software Requirements}
\label{app:env-software}
A minimal reproduction environment includes:
\begin{enumerate}[leftmargin=*]
    \item Python 3.10 or newer,
    \item NumPy, scikit-learn,
    \item matplotlib for figure regeneration.
\end{enumerate}

\subsection{Hardware Assumptions}
\label{app:env-hardware}
No specialized accelerator hardware is required. CPU-only execution is sufficient for all headline numbers.

\subsection{Storage and Logging}
\label{app:env-storage}
Reproduction assumes access to local temporary storage for controlled execution behavior and persistent decision logs for analysis.

\section{Reproducibility Statement}
\label{app:reproducibility}

\subsection{Dataset License and Usage Terms}
\label{app:license}

The \NEXUS benchmarks consist of synthesized plan templates spanning multiple tool categories (file operations, database writes, network calls, and external APIs). All datasets are released under \textbf{CC BY 4.0} and the trained risk scorer under \textbf{Apache-2.0}. By downloading or using any of these artifacts you agree to the following terms.

\begin{itemize}[leftmargin=*,itemsep=2pt,topsep=2pt]
    \item \textbf{Research use only.} The datasets are released \emph{solely for research on agent safety} (monitoring, evaluation, red-teaming, and policy design). Use for offensive security against systems without explicit authorization, or to train models intended to circumvent safety monitors, is out of scope and not authorized.
    \item \textbf{Dataset cards are authoritative.} Each Hugging Face dataset card documents per-dataset construction details, known limitations, sensitive-content disclosures, and any additional usage constraints. Where this paper and the dataset card disagree, the dataset card governs.
    \item \textbf{Synthetic and curated content.} Sensitive-looking arguments inside the datasets (credentials, paths, tables, URLs) are synthetic test fixtures. They are designed to exercise the argument inspector and rule set; they do not refer to real systems, users, or accounts.
    \item \textbf{Attribution.} If you use the datasets, model, or rule set in published work, please cite this paper and link to the relevant Hugging Face dataset card.
    \item \textbf{Redistribution.} You may redistribute the datasets and derived artifacts under the same CC BY 4.0 terms (or, for the scorer, Apache-2.0), preserving attribution and this license notice.
\end{itemize}

The dataset cards live at the URLs listed in \S\ref{sec:hf-release}; consult each card before downloading.

\subsection{Public Release on Hugging Face}
\label{sec:hf-release}

All artifacts are publicly hosted on Hugging Face under the \texttt{EliasHossain} namespace
and are accessible at the following permanent URLs:
\begin{itemize}[leftmargin=*,nosep]
    \item \textbf{Trained model:} \url{https://huggingface.co/EliasHossain/nexus-risk-scorer}
    \item \textbf{Synthetic benchmark:} \url{https://huggingface.co/datasets/EliasHossain/nexus-synthetic}
    \item \textbf{Multi-turn sessions:} \url{https://huggingface.co/datasets/EliasHossain/nexus-multistep}
    \item \textbf{IPI adversarial:} \url{https://huggingface.co/datasets/EliasHossain/nexus-ipi}
    \item \textbf{Stress benchmark:} \url{https://huggingface.co/datasets/EliasHossain/nexus-stress}
\end{itemize}
\noindent Source code and reproduction pipeline: \url{https://github.com/eliashossain001/nexus}

\noindent Table~\ref{tab:hf_releases} summarises the contents of each release.

\begin{table}[h]
\centering
\small
\caption{Hugging Face release inventory.}
\label{tab:hf_releases}
\begin{tabular}{p{2.8cm}llp{5.2cm}}
\toprule
\rowcolor{tabheader}
\textcolor{tabheadertext}{\textbf{Repository}} &
\textcolor{tabheadertext}{\textbf{Type}} &
\textcolor{tabheadertext}{\textbf{License}} &
\textcolor{tabheadertext}{\textbf{Contents}} \\
\midrule
\rowcolor{tabalt}
\texttt{nexus-risk-scorer} & Model & Apache-2.0
  & Calibrated 9-D logistic regression (\texttt{risk\_scorer.pkl}), FAISS index, RAG chunks; 1.3\,KB scorer footprint \\
\texttt{nexus-synthetic} & Dataset & CC BY 4.0
  & 428 structured plans (9 risk categories); JSON, 705\,kB; auto-converted to Parquet \\
\rowcolor{tabalt}
\texttt{nexus-multistep} & Dataset & CC BY 4.0
  & 120 multi-turn sessions (360 turns, 5 risk categories); JSON, 616\,kB \\
\texttt{nexus-ipi} & Dataset & CC BY 4.0
  & 400 paired IPI instances (v1: 200, v2: 200); 5 injection styles in v2; 451\,kB \\
\rowcolor{tabalt}
\texttt{nexus-stress} & Dataset & CC BY 4.0
  & 200 rule-blind adversarial plans; 20 categories $\times$ 11 domains; JSONL+JSON, 668\,kB \\
\bottomrule
\end{tabular}
\end{table}

\paragraph{Model: nexus-risk-scorer.}
The model repository contains three files: (i)~\texttt{risk\_scorer.pkl}, a pickled dictionary
with keys \texttt{model} (scikit-learn \texttt{LogisticRegression}, \texttt{class\_weight='balanced'},
\texttt{max\_iter=1000}, \texttt{random\_state=42}) and \texttt{scaler} (\texttt{StandardScaler});
(ii)~\texttt{faiss.index}, a pre-built FAISS vector index for the SensitiveRegistry; and
(iii)~\texttt{chunks.pkl}, the corresponding RAG knowledge-base chunks.
The model card documents the full 9-D feature schema, intervention policy $\Pi$, threshold 
derivation on a $5 \times 5$ $(\lambda_s, \lambda_o)$ grid, calibration metrics 
(ECE\,=\,0.013, Brier\,=\,0.041), performance across all seven evaluation settings with 
bootstrap CIs, and known limitations including the CONFIRM/REVISE coverage gap.

\paragraph{Dataset: nexus-synthetic.}
Ships as a single JSON array (\texttt{full\_synthetic.json}, 705\,kB, 428 records). 
Each record includes a natural-language request, a structured multi-step plan with per-tool 
side-effect and permission metadata, binary safety labels, 4-class intervention labels 
(\texttt{allow}: 82, \texttt{block}: 235, \texttt{request\_confirmation}: 50, 
\texttt{request\_revision}: 60), and one of 9 risk-type annotations.
Construction: 420 deterministic templates (\texttt{seed=0}) plus 8 harvested real plans.
Splits are reconstructed deterministically via 
\texttt{src/runtime\_safety/benchmarks/split.py}: \texttt{seed=42} for train/test (300/128), 
\texttt{seed=7} for train/calibration (240/60).

\paragraph{Dataset: nexus-multistep.}
Ships as \texttt{multistep\_sessions\_full.json} (616\,kB, 120 sessions).
Each session contains 2--4 turns, each with its own structured plan identical in schema 
to \texttt{nexus-synthetic}.
A \texttt{critical\_turn\_idx} annotation marks the turn at which the cumulative session 
becomes unsafe (\texttt{null} for the 25 legitimate-multi-turn controls).
Five session-level risk categories: cross-step exfiltration (35), 
legitimate multi-turn controls (25), repeated unsafe intent (20), escalating privilege (20), 
and incremental sensitive access (20).
A per-turn-only baseline scores F$_1 \approx 0.5$; session memory achieves 95/95 critical 
turns caught and 25/25 controls allowed (F$_1$\,=\,1.0).

\paragraph{Dataset: nexus-ipi.}
Ships as two Hugging Face configs (\texttt{v1}, \texttt{v2}), each with a \texttt{test} split:
\begin{itemize}[leftmargin=*,nosep]
    \item \textbf{v1} (200 records): 100 adversarial + 100 matched controls using a single 
          canonical injection template. F$_1$\,=\,0.995 [0.98, 1.00].
    \item \textbf{v2} (200 records): 5 injection styles $\times$ 20 pairs: \textit{polite\_request},
          \textit{urgent\_authority}, \textit{social\_engineering}, 
          \textit{code\_fenced\_instruction}, \textit{long\_context\_payload}. F$_1$\,=\,1.000.
\end{itemize}
Each adversarial/control pair shares an identical user request and legitimate first-step 
tool call; only the \texttt{context.history} field differs (clean vs.\ injected), 
isolating sensitivity to injection payloads carried through retrieved content or tool output.
Deterministic generation with \texttt{seed=0}.

\paragraph{Dataset: nexus-stress.}
Ships as both \texttt{nexus\_stress.jsonl} and \texttt{nexus\_stress.json} (668\,kB, 200 records).
Rule-blind adversarial plans authored without matching to the NEXUS rule-set vocabulary, 
spanning 20 scenario categories across 11 domains.
Gold labels cover all four interventions (BLOCK: 71, CONFIRM: 54, REVISE: 40, ALLOW: 35).
Difficulty distribution: medium (64), hard (56), ambiguous (48), easy (25), borderline (7).
91/200 examples are flagged \texttt{is\_intentionally\_difficult\_edge\_case}; 
35/200 are \texttt{is\_safe\_but\_suspicious\_control}.
Semantic deduplication enforces TF-IDF cosine $< 0.80$ between any retained pair; 
deterministic build with \texttt{seed=0}.

\paragraph{Programmatic access.}
All datasets are loadable via the Hugging Face \texttt{datasets} library:
\begin{verbatim}
from datasets import load_dataset
synth  = load_dataset("EliasHossain/nexus-synthetic", split="full")
multi  = load_dataset("EliasHossain/nexus-multistep", split="test")
ipi_v1 = load_dataset("EliasHossain/nexus-ipi", "v1", split="test")
ipi_v2 = load_dataset("EliasHossain/nexus-ipi", "v2", split="test")
stress = load_dataset("EliasHossain/nexus-stress", split="test")
\end{verbatim}
The trained scorer can be loaded directly without cloning the source repository:
\begin{verbatim}
from huggingface_hub import hf_hub_download
import pickle, numpy as np

path = hf_hub_download("EliasHossain/nexus-risk-scorer",
                       "risk_scorer.pkl")
ckpt = pickle.load(open(path, "rb"))
model, scaler = ckpt["model"], ckpt["scaler"]

# Example: 9-D feature vector for a plan
features = np.array([[3, 1, 1, 0, 3, 2, 4.0, 0.08, 0]])
rho = model.predict_proba(scaler.transform(features))[0, 1]
print(f"rho = {rho:.3f}")  # risk score in [0, 1]
\end{verbatim}

\subsection{Software and Hardware Requirements}

\NEXUS is intentionally lightweight so that runtime safety monitoring imposes no infrastructure cost on the agent loop. Table~\ref{tab:environment} lists the full software stack and hardware envelope. The full pipeline (Steps~0--7 in Table~\ref{tab:pipeline}) is pure-CPU, fits in under $2$\,GB of RAM, requires under $50$\,MB of disk, and uses only standard scientific-Python dependencies (\texttt{scikit-learn}, \texttt{NumPy}, \texttt{SciPy}) plus \texttt{faiss-cpu} for the SensitiveRegistry index and \texttt{pydantic} for plan validation. A single CPU core is sufficient for end-to-end reproduction; a $4$-core machine completes Steps~0--7 in roughly four minutes. The only external dependency is an OpenAI API key, and only for Step~8 (the optional GPT-4o LLM-judge comparison in Table~\ref{tab:llm-judge}); skipping Step~8 removes all third-party API requirements.

\begin{table}[h]
\centering
\small
\caption{Environment specification.}
\label{tab:environment}
\begin{tabular}{ll}
\toprule
\rowcolor{tabheader}
\textcolor{tabheadertext}{\textbf{Requirement}} &
\textcolor{tabheadertext}{\textbf{Specification}} \\
\midrule
\rowcolor{tabalt}
Python & $\geq$ 3.10 \\
scikit-learn & $\geq$ 1.3 \\
\rowcolor{tabalt}
NumPy & $\geq$ 1.24 \\
faiss-cpu & $\geq$ 1.7.4 \\
\rowcolor{tabalt}
pydantic & $\geq$ 2.0 \\
SciPy & $\geq$ 1.11 \\
\midrule
\rowcolor{taboos}
GPU & Not required (Steps 0--7) \\
\rowcolor{taboos}
CPU & Single core sufficient; 4-core $\approx$ 4 min \\
\rowcolor{taboos}
RAM & $<$ 2 GB peak \\
\rowcolor{taboos}
Disk & $<$ 50 MB total \\
\rowcolor{taboos}
Step 8 only & OpenAI API key (GPT-4o) \\
\bottomrule
\end{tabular}
\end{table}

\subsection{Model Training and Calibration}

The risk scorer is a logistic regression classifier over the 9-dimensional feature vector 
$\phi(P)$ (Appendix~\ref{app:features}), with Platt calibration (ECE\,=\,0.013)

\begin{table}[h]
\centering
\small
\caption{Data splits and random seeds.}
\label{tab:splits}
\begin{tabular}{llrl}
\toprule
\rowcolor{tabheader}
\textcolor{tabheadertext}{\textbf{Split / Operation}} &
\textcolor{tabheadertext}{\textbf{Purpose}} &
\textcolor{tabheadertext}{\textbf{Size}} &
\textcolor{tabheadertext}{\textbf{Seed}} \\
\midrule
\rowcolor{tabalt}
Stratified test+val split & Evaluation & 191 (test+val) & 42 \\
\rowcolor{taboos}
\quad -- Test set & Headline metrics & 128 & \textcolor{catmot}{--} \\
\rowcolor{taboos}
\quad -- Validation set & Threshold selection & 63 & \textcolor{catmot}{--} \\
\rowcolor{tabalt}
Training set & Model fitting & 300 & \textcolor{catmot}{--} \\
\rowcolor{taboos}
\quad -- Base-fit (80\%) & LR training & 240 & 7 \\
\rowcolor{taboos}
\quad -- Calibration (20\%) & Platt scaling & 60 & 7 \\
IPI regeneration & Adversarial pairs & 200 & 0 \\
\rowcolor{tabalt}
Bootstrap CIs & Confidence intervals & 1{,}000 resamples & 0 \\
Permutation test & 4-class significance & 5{,}000 permutations & 42 \\
\rowcolor{tabalt}
R-Judge 5-fold CV & External retrain & 564 records & 7 \\
\bottomrule
\end{tabular}
\end{table}

\noindent\textbf{Threshold selection.}
Intervention thresholds $(\tau_b, \tau_c) = (0.75, 0.70)$ are selected on the 
63-instance validation split and reported exclusively on the 128-instance test split.
The scorer-gated intervention policy follows a 7-step deterministic cascade 
(Algorithm~1, Appendix~\ref{app:algorithm}).

\subsection{Reproduction Pipeline}

Table~\ref{tab:pipeline} describes the full reproduction pipeline executed by a 
single command.

\begin{table}[h]
\centering
\small
\caption{Reproduction pipeline steps.}
\label{tab:pipeline}
\begin{tabular}{clll}
\toprule
\rowcolor{tabheader}
\textcolor{tabheadertext}{\textbf{Step}} &
\textcolor{tabheadertext}{\textbf{Name}} &
\textcolor{tabheadertext}{\textbf{Output}} &
\textcolor{tabheadertext}{\textbf{Approx. Time}} \\
\midrule
\rowcolor{tabalt}
0 & Validate environment & Dependency check & $<$1s \\
1 & Load benchmarks & Data integrity check & $<$1s \\
\rowcolor{tabalt}
2 & Train risk scorer & \texttt{risk\_scorer.pkl} & $\sim$2s \\
3 & Evaluate synthetic & Table 3 metrics (F1, IntAcc) & $\sim$5s \\
\rowcolor{tabalt}
4 & Evaluate IPI & IPI F1, FPR & $\sim$3s \\
5 & Evaluate sessions & Multi-turn session metrics & $\sim$10s \\
\rowcolor{tabalt}
6 & Evaluate NEXUS-Stress & Stress-test F1 & $\sim$8s \\
7 & Runtime \& statistics & Latency, bootstrap CIs & $\sim$30s \\
\rowcolor{taboos}
8 & \textcolor{catmot}{LLM-judge (optional)} & \textcolor{catmot}{GPT-4o comparison} & \textcolor{catmot}{$\sim$3 min} \\
\bottomrule
\end{tabular}
\end{table}

\begin{verbatim}
# Full reproduction (Steps 0-8):
pip install -e .
python run_complete/run_all_experiments.py

# Without LLM-judge (Steps 0-7, no API cost):
python run_complete/run_all_experiments.py --skip-llm-judge
\end{verbatim}

\subsection{Evaluation Protocol}

\begin{itemize}[leftmargin=*]
    \item \textbf{Primary metrics} (Table 3): Binary F1, Intervention Accuracy, 
          per-class precision/recall, all on the held-out 128-instance test split.
    \item \textbf{Bootstrap CIs}: 1{,}000 stratified resamples (seed=0); 
          95\% percentile intervals.
    \item \textbf{Statistical tests}: Exact McNemar (binary); paired permutation test 
          (4-class, 5{,}000 permutations, seed=42).
    \item \textbf{R-Judge}: 564 usable records (7 leaked excluded from 571); 
          tool metadata inferred via keyword tables; 5-fold CV retrain (seed=7), 
          majority-vote baseline.
    \item \textbf{NEXUS-Stress}: 200 instances, 20 categories $\times$ 11 domains; 
          TF-IDF $\geq 0.80$ duplicate rejection at generation time.
    \item \textbf{IPI}: 200 matched safe/injected pairs per version; deterministic seed=0 
          generation; v2 ablates 5 injection styles.
    \item \textbf{Sessions}: 120 multi-turn sessions; session-level accuracy 
          (95/95 true positives, 25/25 controls allowed).
    \item \textbf{Latency}: Median over 384 evaluations 
          (3 runs $\times$ 128 instances), single CPU core, no batching.
    \item \textbf{AgentHarm}: 44 benign + 66 harmful behaviors; evaluated against 
          GPT-4o safety baseline.
\end{itemize}

\subsection{Determinism Guarantees}

All experiments (Steps 0--7) are fully deterministic on CPU given the specified seeds. 
Step~8 (GPT-4o comparison) depends on the OpenAI API and may exhibit minor 
non-determinism across runs due to model serving variability. 
Results are bit-reproducible for all non-LLM experiments.

\section{Ethics Statement}
\label{app:ethics}
The synthetic benchmark contains no personally-identifiable information. The IPI adversarial set uses fabricated credentials (e.g.\ the dummy API key \texttt{sk\_live\_4xKj9...}) and synthetic URLs (\texttt{attacker.example.com}). The IPI templates we release can in principle be used to probe deployed agents; we follow the release norm of \citet{yuan2024r} and \citet{andriushchenko2025agentharm} in that surfacing classes of failure benefits defenders more than withholding them benefits attackers. \NEXUS is positioned for use in evaluation and as one component of a deployed safety stack; we explicitly do not claim that it provides full assurance against adversaries with access to the agent's LLM weights, the tool implementations, or covert side channels (Table~\ref{tab:threats}).

\section{AI Usage Statement}
\label{app:ai-usage}
Generative AI assistants were used to support drafting and editing of the manuscript and to accelerate routine code-development tasks, such as writing docstrings, adding type hints, and performing minor refactors. All technical claims, experimental designs, empirical results, ablation choices, and figures were defined, executed, verified, and finalized by the authors. No AI assistant was used to fabricate, alter, or synthesize experimental results; every reported value is computed from the released code and benchmarks. The trained risk scorer, rule set, argument inspector, IPI templates, and multi-turn benchmark were authored and validated by the authors.

\end{document}